\documentclass{article}
\usepackage{amsmath} 
 \usepackage{graphicx}
  \usepackage{amssymb}
  \PassOptionsToPackage{numbers, compress}{natbib}
\usepackage{booktabs}
\usepackage{makecell}

\usepackage[preprint]{neurips_2026}
\usepackage[utf8]{inputenc} 
\usepackage[T1]{fontenc}    
\usepackage{hyperref}       
\usepackage{url}            
\usepackage{booktabs}       
\usepackage{amsfonts}       
\usepackage{nicefrac}       
\usepackage{microtype}      
\usepackage{xcolor}         
\usepackage{natbib}
\usepackage{enumitem} 
\usepackage{algorithm}
\usepackage{multirow}
\usepackage{algorithmic}

\title{Neural Structural Reasoner: A Brain-inspired Architecture for Reasoning over Structured Knowledge}

\author{%
  Zixing Jia$^{1,2,3,4}$\thanks{Equal contribution.}
  \qquad
  Yuhang Pan$^{1,2,3}$\footnotemark[1]
  \qquad
  Ni Ji$^{1,2,3}$\thanks{Corresponding author.} \\[4pt]
  $^{1}$Beijing Institute for Brain Research, Chinese Academy of Medical Sciences \\ \& Peking Union Medical College, Beijing, 102206, China \\
  $^{2}$Chinese Institute for Brain Research, Beijing; Beijing, 102206, China \\
  $^{3}$Beijing Key Laboratory of Brain Science and Brain-Machine Interface \\
  $^{4}$Sun Yat-sen University, Guangzhou, China \\[3pt]
  \texttt{J3540493668@outlook.com}
  \quad
  \texttt{panyuhang@cibr.ac.cn}
  \quad
  \texttt{niji@cibr.ac.cn}
}

\begin{document}

\maketitle

\begin{abstract}
  Structural reasoning, the ability to recognize and make inferences over the relational structure between objects and concepts, is a hallmark of human cognition, yet prevailing methods often collapse relational topology into flat embeddings, cannot discover hidden structure and lack interpretability. We introduce Neural Structural Reasoner (NSR), a brain-inspired network that preserves relational structure directly in the connectivity and dynamics of coupled neuronal populations. NSR draws inspiration from three biological mechanisms: multi-layered architecture for encoding hierarchical knowledge, stable representations of entity and concepts, and path integration for input-driven state inference. At query time, NSR parallelizes computation over candidate relational structures and leverages confidence-weighted scores to perform link prediction. Across standard knowledge-graph benchmarks, NSR achieves competitive accuracy without leading on every dataset, and has lower reported training times than several neural baselines. Because reasoning is implemented through sequences of human-readable neuron activations, NSR affords native interpretability by tracking intermediate inference steps. The model further extracts latent relational hierarchies and compositional rules, demonstrating the brain-inspired architecture as an effective, efficient, and highly interpretable substrate for structural reasoning.
\end{abstract}

\section{Introduction}
Knowledge in the real world is inherently structured—organized by relations such as order, hierarchy, and connectivity—and humans routinely rely on these relations, rather than the entities alone, to reason about the world \cite{fodor1988,gentner1983}. This ability to understand and reason about data relationships, which we call \emph{structural reasoning}, has long been regarded as a desideratum for human-like machine intelligence \cite{brenden2017,webb2024}. Machine learning pursues it from several angles: knowledge-graph embeddings map entities and typed relations into continuous spaces for multi-hop queries \cite{hogan2021,ji2022,wang2017};graph neural networks impose relational inductive biases via message passing \cite{battaglia2018,santoro2017,petar2022};  and large language models, now widely deployed as general reasoners, tackle structural tasks by verbalizing relations and chaining inferences in natural language \cite{wu2023,he2026,dziri2023,press2023}. Despite their differences, these methods share two recurring weaknesses: they tend to collapse rich structural information into flat vector representations and produce answers without exposing how they were reached—making their reasoning hard to verify and prone to break on deeper, multi-step queries \cite{barrett2018,zhang2020}.

Biology offers an alternative starting point. Structural reasoning in the brain arises from the coordinated activity of several systems: the hippocampal–entorhinal circuit builds cognitive maps—relational organizations of entities that support inference beyond direct experience across spatial and non-spatial domains \cite{behrens2018,whittington2020}, and does so predictively and in a manner causally required for structure-based inference \cite{stachenfeld2017,pan2025causal}; the prefrontal cortex abstracts relations into hierarchical, schema-like representations that generalize across contexts \cite{samborska2022complementary}.Existing biologically inspired models, including the Tolman–Eichenbaum Machine \cite{whittington2020}, Vector-HaSH \cite{chandra2025episodic}, and related associative-memory architectures \cite{steinberg2022associative,tahir2024long,ramsauer2021hopfield}, draw on these neural principles to represent and retrieve structured knowledge. However, to our knowledge, there still remains a lack of a modular, brain-inspired architecture that can both encode and perform reasoning over structured knowledge at scale.

We address this gap with Neural Structural Reasoner (NSR), a brain-inspired network architecture for structural reasoning.  The main advances are four-fold:
\begin{itemize}[leftmargin=*, topsep=2pt, itemsep=2pt, parsep=0pt]
    \item \textbf{Network structure as relational structure.} NSR uses network  connectivity to represent relational structure, and exploits efficient network dynamics to implement relational reasoning.
    \item \textbf{Parallel reasoning with path-support scores.} NSR computes in parallel over multiple candidate relational structures, returning ranked candidate answers with confidence score.
    \item \textbf{Competitive performance with high computational efficiency.} NSR achieves competitive, dataset-dependent accuracy across benchmarks while requiring relatively less training time.
    \item \textbf{Interpretability from dynamics.} NSR yields human-readable reasoning paths by tracking network dynamics as reasoning steps at inference.
\end{itemize}

\section{Background and Related Works}
\textbf{2.1 Structural reasoning across machine learning}

Structural reasoning has long been studied under Knowledge Representation and Reasoning (KRR)~\cite{brachman2004knowledge}, which formalizes knowledge through structured objects---logic rules, ontologies, and graphs---and defines inference as a process operating on that structure. Among these, knowledge graphs (KGs) offer an instantiation that is both scalable in practice and conducive to systematic empirical study. Formally, a KG is a tuple $\mathcal{G} = (\mathcal{E}, \mathcal{R}, \mathcal{T})$ with entities $\mathcal{E}$, relations $\mathcal{R}$, and triples $\mathcal{T} \subseteq \mathcal{E} \times \mathcal{R} \times \mathcal{E}$, where each $(h, r, t) \in \mathcal{T}$ asserts that entity $h$ stands in relation $r$ to entity $t$ \cite{hogan2021,ji2022}. The canonical task is link prediction: given a query $(h,r,?)$, score candidate tails and recover the missing entity from the relational structure already in $\mathcal{G}$. This makes structural reasoning concrete---inference over typed relations among discrete entities, isolated from perceptual or linguistic confounds.

A central object of study is the structure of relations themselves, since real KGs are pervaded by canonical patterns like symmetry (e.g., \texttt{sibling\_of}), inversion ($r_1(x,y) \Rightarrow r_2(y,x)$, e.g., \texttt{teacher\_of} $\leftrightarrow$ \texttt{student\_of}), and composition ($r_1(x,y) \land r_2(y,z) \Rightarrow r_3(x,z)$, e.g., \texttt{father\_of} $\circ$ \texttt{father\_of} $\Rightarrow$ \texttt{grandfather\_of})~\cite{sun2018rotate}. Composition is especially important: nearly every relation in benchmark KGs participates in some compositional pattern~\cite{niu2024knowledge}. Existing methods address these regularities through distinct representational strategies, each with characteristic limitations. \textbf{Geometric embedding and tensor-factorization models} encode relations as translations, rotations, bilinear maps, or low-rank interactions, enabling scalable link prediction but leaving composition implicit in vector geometry~\cite{bordes2013translating,sun2018rotate, nickel2011three, trouillon2016complex}. \textbf{Deep neural models}, including ConvE~\cite{dettmers2018convolutional} and message-passing GNNs~\cite{schlichtkrull2018modeling, zhu2021neural, galkin2023towards}, add nonlinear computation and neighborhood aggregation, but often lack transparent inference trajectories and struggle with long-range relational chains. \textbf{Rule-mining and path-based systems} expose logical chains, but rely on combinatorial search and are brittle under missing or noisy edges~\cite{yang2017differentiable, das2017go}. LLM-based methods can verbalize triples or retrieve subgraphs, yet their outputs are prompt-sensitive and may not faithfully track executable graph operations. Thus, current methods generally lack a native mechanism for representing relational structure as explicit, reusable, and traceable dynamics over a structured network.

\textbf{2.2 Neural substrates of structural reasoning}

Structural reasoning in the brain appears to rely on a distributed network of brain regions. Conceptual knowledge is often linked to the anterior temporal lobe (ATL), a proposed transmodal semantic hub that integrates modality-specific features into \textbf{stable, context-general concepts}~\cite{patterson2007you, ralph2017neural}. Relational operations recruit prefrontal mechanisms, especially rostrolateral/frontopolar prefrontal cortex, whose \textbf{activity increases with relational integration} demands and which is implicated in analogical and multi-relational reasoning~\cite{christoff2001rostrolateral, bunge2005analogical, wendelken2008brain, waltz1999system}. \textbf{Flexible composition over relations} further depends on medial temporal lobe circuitry: the hippocampus binds overlapping experiences into relational structures that support novel inference~\citep{dusek1997hippocampus, preston2004hippocampal, eichenbaum2004conditioning}, while entorhinal grid-like codes may provide a metric format for cognitive maps spanning physical and conceptual spaces~\citep{constantinescu2016organizing, park2021inferences}. Through \textbf{path integration}, hippocampal–entorhinal circuits update an internal state estimate by accumulating transitions through space~\cite{behrens2018}; by extension, analogous dynamics may support abstract reasoning by tracking trajectories through relational or conceptual spaces (though direct evidence remains scarce). Thus, structural reasoning in the brain likely emerges from specialized but interconnected regions that support stable entity and relation representations, relation-specific binding of entities, and compositional generalization of relational structures.

\textbf{2.3 Biologically inspired models of structural reasoning} 

Some biologically inspired architectures offer important ingredients for relational cognition, though they have not yet fully addressed structural reasoning in the sense of §2.1. The Tolman–Eichenbaum Machine (TEM)~\cite{whittington2020,whittington2021relating} repurposes hippocampal–entorhinal path integration for abstract inference, treating relations as velocity-like signals over a learned structural scaffold; however, its entorhinal module is implemented as a backpropagation-trained RNN, leaving open how such computations arise from grid-cell-like dynamics. Vector–HaSH~\cite{chandra2025episodic} provides a more biologically grounded HPC–EC loop and supports sequence memory via path integration, but its role in abstract relational reasoning remains unexplored. Continuous-attractor models of grid cells~\cite{burak2009accurate} offer a mechanistic account in which velocity-related inputs drive state transition to encode movement in space, yet their application is often limited to low-dimensional navigation rather than high-dimensional relational domains. A deeper limitation of many HPC-EC inspired models is that the network dynamics, built for 2D navigation, are confined to a 2D torus, while \textbf{abstract relations require input-driven transitions on high-dimensional manifolds}—and no biologically plausible architecture have been shown to supports this. Together, these works motivate a bio-inspired architecture that can encode and perform reasoning in structured, high-dimensional relation space.

\section{The Neural Structural Reasoner Model}
\label{sec:nsr-model}

\begin{figure}[htbp]

     \centering
        \includegraphics[width=0.9\linewidth]{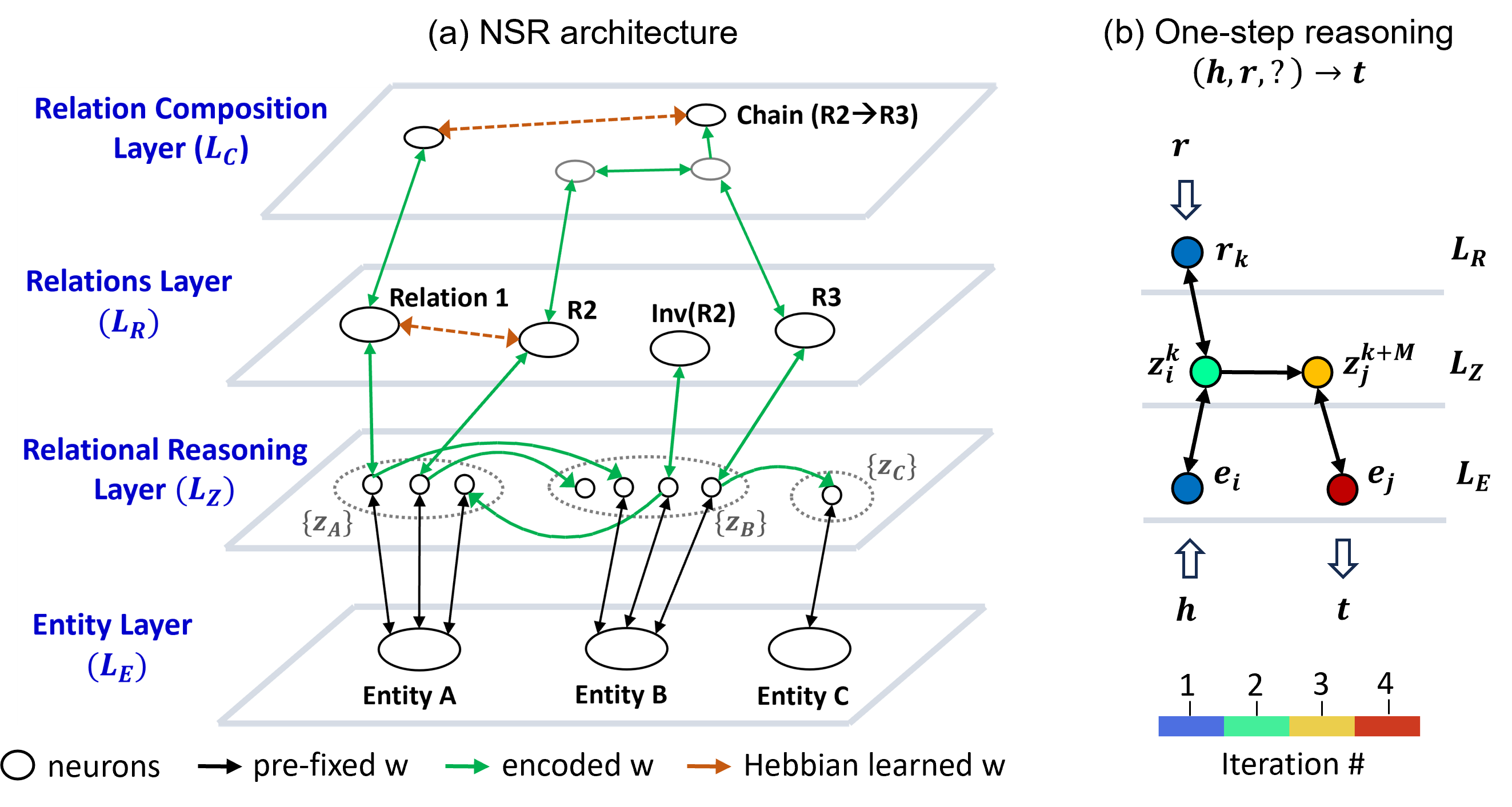}

    \caption{\textbf{The Neural Structural Reasoner Architecture.} (a) Four core layers of NSR: $L_E$ encode entities; $L_Z$ instantiates relation-driven state transition;  $L_R$ and $L_C$ encode relations and their compositions (e.g., chains). Connections in black are fixed a priori, those in green encode observed triples, those in orange are learned via Hebbian updates to capture latent relational structure. (b) Sequence of neuron activation that implements one-step reasoning over $(h, r, \, ?) \mapsto t$.}

    \label{fig:NSR_Framework}
\end{figure}

\subsection{Model Architecture}
\label{sec:model-architecture}

Neural Structural Reasoner (NSR) maps relational structure between entities onto a hierarchical multi-layered network, where connectivity directly encodes associations between entities, relations and relation compositions (Fig. \ref{fig:NSR_Framework}a). The core functional layers are:

\textbf{Entity Layer ($L_E$)}: contains neurons $\{e_i\}^N$ encoding the $N$ entities in the data.

\textbf{Relations Layer ($L_R$)}: contains a total of $2M$ neurons where the first half, $\{r_i\}^{M}$, encode the $M$ relations and the second half, $\{r_{i+M}\}^{M}$, encode the inverse relations $\{r^{-1}\}^{M}$ (e.g., \texttt{father\_of}$^{-1}$ is learned to associate with \texttt{son\_of} via the Hebbian rule in \S3.3).

\textbf{Relational Reasoning Layer ($L_{Z}$)}: contains up to $N \times (2M)$ neurons; a given neuron $z_i^k$ encodes the association between the head entity $h_i$ and the relation $r_k$. 

\textbf{Relation Composition Layer ($L_{C}$)}: contains a flexible number of neurons encoding the compositional structures (e.g. chains) among relations.

We assess the functional contribution of the mechanisms implemented by these layers in Appendix~\ref{app:ablation}, through ablations of inverse-relation encoding, relation-equivalence retrieval, compositional inference, and Hebbian updates.

\subsubsection{Network Connectivity and Dynamics} The four layers are connected in a bi-directional hierarchy: $L_E \leftrightarrow L_Z \leftrightarrow L_R \leftrightarrow L_C$. $L_E$ and $L_Z$ neurons encoding the same entity are bidirectionally coupled at initialization and held fixed: 

{
\setlength{\abovedisplayskip}{4pt}
\setlength{\belowdisplayskip}{4pt}
\setlength{\abovedisplayshortskip}{2pt}
\setlength{\belowdisplayshortskip}{2pt}
\begin{equation}
\forall i \in [N],\;\forall k \in [2M]:\quad
W_{E\leftrightarrow Z}(e_i, z_i^k) = 1
\label{eq:weightinit}
\end{equation}
}

All other weights are initialized at 0 and are learnable.  The activation state of each layer is updated in a single step by summing up intra- and inter-layer inputs and external query input $I$, and transformed through sigmoidal nonlinearity $\sigma$.

\begin{equation}
\left\{
\begin{aligned}
L_{E}: X_E(t+1) &= \sigma_E\!\left(W_{E\leftrightarrow Z}X_Z(t)\right)+I_E, \\
L_{Z}: X_Z(t+1) &= \sigma_Z\!\left(W_{ZZ}H\bigl(X_Z(t)-\eta_Z\bigr)
+ W_{Z\leftrightarrow E}X_E(t)
+ W_{Z\leftrightarrow R}H\bigl(X_R(t)-\eta_R \bigr) \right)+I_Z, \\
L_{R}: X_R(t+1) &= \sigma_R\!\left(W_{RR}X_R(t)  
+ W_{R\leftrightarrow Z}H\bigl(X_Z(t)-\eta_Z\bigr)
+ W_{R\leftrightarrow C}X_C(t)\right)+I_R, \\
L_{C}: X_C(t+1) &= \sigma_C\!\left(W_{CC}X_C(t)
+ W_{C\leftrightarrow R}H\bigl(X_R(t)-\eta_R\bigr)\right)+I_C
\end{aligned}
\right.
\label{eq:fulldyn}
\end{equation}

Neuronal interaction within $L_Z$ are rectified by the Heaviside function $H$, ensuring that activity propagates to only in response to convergent inputs from the associated entity and relation. Similar gating exists for projections $L_Z\rightarrow L_R$ and $L_R\rightarrow L_C$ to conditional activation of co-occurring relations and relation compositions in the learning phase (§3.3). When performing reasoning, the relevant entities and relations obtained in intermediate steps can be read out directly from neuronal activations in $L_E$, $L_R$ or $L_C$. 

\subsection{Encoding of Relation Triples}
For each triple $(h_i, r_k, t_j)$ and its inverse $(t_j, r_{k}^{-1}, h_i)$, we let neurons $r_k$ and $r_{k+M}$ to encode the relation and its inverse, and connect them with the entity-encoding neurons by assigning the following weights :

\begin{equation}
\left\{
\begin{aligned}
& W_{ZZ}(z_i^k, z_j^{k+M}) = W_{ZZ}(z_j^{k+M}, z_i^k) = 1, \\
& W_{Z\leftrightarrow R}(z_i^k, r_k) = 1, \quad
  W_{Z\leftrightarrow R}(z_j^{k+M}, r_{k+M}) = 1
\end{aligned}
\right.
\label{eq:encw}
\end{equation}

\subsubsection{Reasoning over $(h_i,r_k,?)$ through Path-Integration-Inspired Dynamics} When queried with the relation triple $(h_i,r_k,?)$ (Fig. \ref{fig:NSR_Framework}b), NSR first activates the entity neuron $e_i$ and the relation neuron $r_k$. Iterating over \eqref{eq:fulldyn} then activates neurons $z_i^k$ and $z_j^{k+M}$ through intra-layer dynamics in $L_Z$. This in turn activates neuron $e_j$, which allows the readout of the tail entity $t_j$. This computational logic is inspired by path integration models proposed for entorhinal grid cells \citep{burak2009accurate,chandra2025episodic}.

\subsection{Learning Compositional Rules among Relations}
\label{sec:learncomp_main}
Following the encoding phase, NSR uses Hebbian-like associative learning to extract the following compositional structure and equivalence rules among relations:

\textbf{Relational Equivalence}:  When queried with two relation triples with the same head entity, e.g. $(h_i,r_k,t_j)$ and $(h_i,r_l,t_w)$, $L_Z$ neurons encoding the tail entities, $z_j^{k+M}$ and $z_w^{l+M}$ will be activated. If in fact $t_j=t_w$, then the activity of $z_j^{k+M}$ and $z_j^{l+M}$ can be elevated above the threshold $\eta_z$ through mutual coupling with $e_j$. Iterating over \eqref{eq:fulldyn} leads to co-activation of $r_{k+M}$ and $r_{l+M}$ and of $r_{k}$ and $r_{l}$. The connections between co-activated relation neurons are updated via a symmetrized version of Oja's rule:

\begin{equation}
\Delta w_{ij}^{RR} =  \delta_{RR}(r_i*r_j)-\mu_{RR}(r_i^2+r_j^2)w_{ij}^{RR} 
\label{eq:hebb}
\end{equation}

Where $\delta_{RR}$ and $\mu_{RR}$ are learning rates. In practice, all relation triples with the same head entity can be queried at once, and weight update can occur simultaneously for all co-activating $r_k$s. This ultimately leads to strong coupling between neurons encoding semantically analogous relations.  

\textbf{Symmetric and Inverse Relations}: Same network dynamics and learning rule can be applied to extract symmetric and inverse relations. This is achieved by letting $r_l=r_l^{-1}$ in the case of symmetry, and  $r_l=r_m^{-1}$ in the case of inverse relations. 






\textbf{Composite relations}: We focus on detecting the equivalence between multi-hop relation chains and single triples. Take two-hop chains for example, this amounts to detecting
\begin{equation}
r_p(h_i,x_p)\land r_q(x_p,x_q)
\Rightarrow r_k(h_i,x_q)
\Leftrightarrow r_k^{-1}(x_q,h_i).
\label{eq:comp-rule}
\end{equation}
where the bidirectional equivalence follows from Eq.~\ref{eq:encw}: every triple $(h,r,t)$ is encoded together with its inverse $(t,r^{-1},h)$. The network discovers such rules by checking whether a two-hop chain returns to the original head entity. Concretely, suppose $(h_i,r_p,x_p)$ is already encoded. NSR computes three inference steps:

\textbf{\textit{Step 1:}} Activate $e_{x_p}$ (the tail of the known triple) and all relation neurons $\{r_m\}_{m=1}^{2M}\setminus\{r_{p+M}\}$ (all relations except the inverse $r_p^{-1}$). Iterating dynamics in $L_Z$ activates neurons $z_q^m$ representing candidate tails $x_q$ reachable from $x_p$ via each relation $r_m$.

\textbf{\textit{Step 2:}} For each discovered entity $x_q$, treat it as a new head entity and repeat Step~1, querying $(x_q,r_m,?)$. This yields a second-hop set of tails $\{x_s\}$ encoded by neurons $z_s^m$.

\textbf{\textit{Step 3:}} If any $x_s$ coincides with the original head entity $h_i$, the network has closed a three-step loop:
\begin{equation}
(h_i,r_p,x_p)\land(x_p,r_q,x_q)
\land(x_q,r_k^{-1},h_i).
\label{eq:three-step}
\end{equation}
Because the loop returns to $h_i$ through $r_k^{-1}$, the two-hop
composition $r_p\circ r_q$ connects $h_i$ to $x_q$ in the direction
of $r_k$, supporting the rule $r_p\circ r_q \Rightarrow r_k$
(Eq.~\ref{eq:comp-rule}). Owing to the bidirectional encoding in
Eq.~\ref{eq:encw}, the presence of $(x_q,r_k^{-1},h_i)$ is equivalent
to the presence of $(h_i,r_k,x_q)$; hence the discovered chain is
registered in $L_C$ as a compositional association between
$r_p\!\to\!r_q$ and $r_k$.

Whenever such a closed loop is detected, assign a $L_C$ neuron $c_k$ to represent the relation $r_k$, and neuron $c_{p\rightarrow q}$ to detect the sequential activation of $r_p\rightarrow r_q$. Update weights according to:

\begin{equation}
\Delta w_{k-pq}^{CC} =  \delta_{CC}(c_k*c_{pq})-\mu_{CC}(c_k^2+c_{pq}^2)w_{k-pq}^{CC} 
\label{eq:hebb-c}
\end{equation}

While these procedures are designed for online sequential learning, for static datasets, the existence of composite structures can be detected statistically and then encoded directly in network weights. We present details of these simplifications along with pseudocodes in Appendix~\ref{sec:compositional-rules}.

\subsection{Reasoning Phase} 
\label{sec:reasoning-phase}
Given a query $(h_i,r_k,?)$ where $(h_i,r_k,t^\ast)\notin \mathcal{D}_{\text{train}}$, NSR uncovers $t^*$ by first retrieving equivalent relations $r_{e}$ or relation compositions $C_{e}$ to $r_k$, and then performing relational reasoning over these equivalence sets to obtain candidate tail entities $\tilde{\mathcal{T}} = \{\tilde{t}\}$ with path-support scores used for ranking:

\textbf{\textit{Step 1} Retrieval of equivalent relations:}
This is achieved by first activating neuron $r_k$ encoding the query relation, and iterating the dynamics of $L_R$ in \eqref{eq:fulldyn} once to activate neurons $\{r_{e}\}$ encoding the set of equivalent relations. A threshold $\texttt{T\_{thresh}}$ gates the propagation, retaining only neurons with activation at least $\texttt{T\_{thresh}}$  ; their activation levels, which represent how strongly two relations are associated, serve as path-support scores (i.e., $S_R(r_e)=x_e$).

\textbf{\textit{Step 2} Relational reasoning with $\{r_{e}\}$:}
This is implemented by activating the neuron $e_i$ encoding the head entity, while holding $\{r_{e}\}$ active. Iterating over the dynamics of $L_Z$ according to \eqref{eq:fulldyn} implements $(h_i,r_e,?)$ and subsequently activates neurons $\{\tilde{e}\}$  in $L_E$. These neurons encode a set of candidate tail entities $\tilde{\mathcal{T}}^{(1)} = \{\tilde{t}\}$.

\textbf{\textit{Step 3} Retrieval of equivalent relation compositions:}
Upon re-initializing the network, activate the query relation neuron $r_k$ and allow its activity to propagate to $L_C$. Iterating $L_C$ dynamics based on \eqref{eq:fulldyn} activates neurons encoding relation compositions that are closely associated with $r_k$. Applying a gate function yields the top-K activated neurons $\{c_{e}\}=\operatorname{TopK}(X_C)$ corresponding to the top equivalent compositions. The activation of $\{c_{e}\}$ serves as the path-support scores $S_C$.

\textbf{\textit{Step 4} Relational reasoning with $\{c_{e}\}$:} Sample and activate a single neuron in $\{c_{e}\}$, which then initiates the sequential activation of $L_R$ neurons $r_{p_1},r_{p_2},\ldots,r_{p_\ell}$ in the relation chain. Activate the entity neuron $e_i$ together with $r_{p_1}$ to initiate multi-hop reasoning in $L_Z$. At the end of the chain traversal, retain the contributing
grounded paths together with their terminal entities in $L_E$.
Each path receives the support score $S_C(c_e)$ of its
corresponding composition neuron.
Repeating this procedure for all selected composition neurons
yields the compositional path collection and its candidate
tail set $\tilde{\mathcal{T}}^{(2)}$.
Distinct groundings of the same relation chain are retained
separately for aggregation in Step~5.

\textbf{\textit{Step 5} Aggregating path-support scores for candidate entities:}
For each candidate tail entity $t_j$, let $\mathcal{P}(t_j)$ be the set of grounded equivalent-relation and compositional paths reaching it, and let $s(p)$ be the support score assigned to path $p$. We compute
\begin{equation}
\mathrm{Score}_{a}(t_j)=
\begin{cases}
\displaystyle\max_{p\in\mathcal{P}(t_j)}s(p), & a=\mathrm{max},\\[4pt]
\displaystyle\sum_{p\in\mathcal{P}(t_j)}s(p), & a=\mathrm{sum},
\end{cases}
\label{eq:score-aggregation}
\end{equation}
where the aggregation mode $a$ is selected on the validation set. Max aggregation may overvalue a single spurious path, whereas sum aggregation may overcount correlated paths. These scores are used to rank candidates, not as calibrated probabilities; the contributing paths remain inspectable.

\section{Experiments}

\subsection{Experimental setup}
\label{app:exp_setup}


\paragraph{Datasets.}
Our main-table evaluation uses four benchmarks: Nations~\cite{kemp2006learning}, Kinship~\cite{brenden2017}, YAGO3-10~\cite{dettmers2018convolutional}, and FB15k-237~\cite{toutanova2015observed}, accessed through PyKEEN~\cite{ali2021pykeen}. These benchmarks span compact relational graphs and larger knowledge graphs. We additionally report experiments on two standard benchmarks, Countries S3~\cite{nickel2016holographic} and WN18RR~\cite{dettmers2018convolutional}, and on our constructed Kinship1990\_EXTENDED dataset in Appendix~\ref{app:additional-benchmarks}. The construction of Kinship1990\_EXTENDED is described in Appendix~\ref{app:kinship-extended}. For the six public benchmarks, all locally evaluated methods use the provided train/validation/test splits.

\paragraph{Evaluation.} We adopt the filtered link-prediction setting \cite{bordes2013translating}, reporting Mean Reciprocal Rank (MRR) and \texttt{Hits@K} ($K \in \{1,3\}$). In addition to predictive accuracy, we measure training time to assess computational efficiency.

\subsection{Empirical performance}
\label{sec:EP}

Tables~\ref{tab:nations_kinship} and~\ref{tab:yago_fb} compare NSR with ten baselines spanning major approaches to knowledge-graph reasoning: embedding models (ConvE~\cite{dettmers2018convolutional} and RotatE~\cite{sun2018rotate}), symbolic rule mining (AnyBURL~\cite{meilicke2024anytime} and AMIE~\cite{galarraga2013amie}), path ranking (PRA/PathRank~\cite{lao2011random}), neural rule learning (NeuralLP~\cite{yang2017differentiable}, NCRL~\cite{cheng2023neural}, and RNNLogic~\cite{qu2021rnnlogic}), reinforcement-learning path search (MINERVA~\cite{das2017go}), and graph neural reasoning (NBFNet~\cite{zhu2021neural}). We report filtered MRR and Hits@1/3 on Nations, Kinship, YAGO3-10, and FB15k-237. Locally evaluated methods use the same dataset splits and filtered-tail evaluation protocol. 
\begin{table*}[t]
\centering
\caption{Link-prediction performance on Nations and Kinship.
Hits@1 and Hits@3 are percentages.
Higher ranking scores and lower training times are better;
the best value in each column is bold.}
\label{tab:nations_kinship}

\small
\setlength{\tabcolsep}{3pt}
\renewcommand{\arraystretch}{1.08}

\resizebox{\textwidth}{!}{%
\begin{tabular}{@{}llcccccccc@{}}
\toprule
& &
\multicolumn{3}{c}{Nations}
& \multicolumn{3}{c}{Kinship}
& \multicolumn{2}{c}{Training time} \\
\cmidrule(lr){3-5}
\cmidrule(lr){6-8}
\cmidrule(lr){9-10}
Category & Method
& MRR & H@1 & H@3
& MRR & H@1 & H@3
& Nations & Kinship \\
\midrule

\multirow{2}{*}{\shortstack[l]{Embedding-\\based}}
& ConvE
& 0.8029 & 69.05 & 89.45
& \textbf{0.7927} & \textbf{68.11} & \textbf{88.45}
& 26 s & 64 s \\

& RotatE
& 0.5351 & 33.23 & 66.02
& 0.7598 & 63.37 & 86.26
& 24 s & 84 s \\

\addlinespace[4pt]

\multirow{2}{*}{\shortstack[l]{Symbolic\\rule learning}}
& AnyBURL
& 0.7994 & 69.15 & 89.55
& 0.6768 & 54.10 & 76.63
& 64 s & 63 s \\

& AMIE
& \textbf{0.8559} & \textbf{77.11} & \textbf{92.54}
& 0.6767 & 55.03 & 76.26
& 0.6 h & \textbf{2 s} \\

\addlinespace[4pt]

\multirow{3}{*}{\shortstack[l]{Neural\\rule learning}}
& NeuralLP
& 0.6841 & 52.74 & 81.59
& 0.6072 & 47.30 & 68.06
& 40 s & 26 s \\

& NCRL
& 0.4571 & 25.37 & 55.22
& 0.6050 & 45.81 & 68.99
& 134 s & 99 s \\

& RNNLogic
& 0.7216 & 61.65 & 83.06
& 0.6690 & 54.94 & 77.60
& 32 s & 82 s \\

\addlinespace[4pt]

\multirow{2}{*}{\shortstack[l]{Path-based\\reasoning}}
& PRA / PathRank
& 0.5933 & 38.31 & 77.61
& 0.6296 & 46.83 & 73.56
& 34 s & 86 s \\

& MINERVA
& 0.5865 & 44.28 & 73.13
& 0.6253 & 46.37 & 73.46
& 0.9 h & 1.2 h \\

\addlinespace[4pt]

GNN-based
& NBFNet
& 0.7479 & 62.19 & 84.08
& 0.7445 & 62.38 & 83.61
& 29 s & 77 s \\

\midrule

Brain-inspired
& NSR
& 0.8142 & 71.64 & 88.16
& 0.6515 & 54.21 & 70.67
& \textbf{3 s} & 11 s \\

\bottomrule
\end{tabular}%
}
\end{table*}

\begin{table*}[t]
\centering
\caption{Link-prediction performance on YAGO3-10 and FB15k-237.
Hits@1 and Hits@3 are percentages.
Higher ranking scores and lower training times are better;
the best available value in each column is bold.
A dash indicates a missing entry.}
\label{tab:yago_fb}

\small
\setlength{\tabcolsep}{3pt}
\renewcommand{\arraystretch}{1.08}

\resizebox{\textwidth}{!}{%
\begin{tabular}{@{}llcccccccc@{}}
\toprule
& &
\multicolumn{3}{c}{YAGO3-10}
& \multicolumn{3}{c}{FB15k-237}
& \multicolumn{2}{c}{Training time} \\
\cmidrule(lr){3-5}
\cmidrule(lr){6-8}
\cmidrule(lr){9-10}
Category & Method
& MRR & H@1 & H@3
& MRR & H@1 & H@3
& YAGO3-10 & FB15k-237 \\
\midrule

\multirow{2}{*}{\shortstack[l]{Embedding-\\based}}
& ConvE
& \textbf{0.6365} & \textbf{59.03} & \textbf{71.28}
& 0.4095 & 31.61 & 44.89
& 10.3 h & 991 s \\

& RotatE
& 0.1812 & 9.98 & 21.04
& 0.3368 & 27.07 & 41.51
& 1.5 h & 1.5 h \\

\addlinespace[4pt]

\multirow{2}{*}{\shortstack[l]{Symbolic\\rule learning}}
& AnyBURL
& 0.5589 & 50.78 & 60.16
& 0.332$^{*}$ & 24.7$^{*}$ & ---
& 1000 s & 1000 s$^{*}$ \\

& AMIE
& 0.5473 & 49.58 & 59.27
& 0.2170 & 16.57 & 23.07
& \textbf{83 s} & \textbf{8 s} \\

\addlinespace[4pt]

\multirow{3}{*}{\shortstack[l]{Neural\\rule learning}}
& NeuralLP
& --- & --- & ---
& 0.3166 & 24.45 & 34.31
& --- & 11.9 h \\

& NCRL
& 0.380$^{*}$ & 27.40$^{*}$ & ---
& 0.300$^{*}$ & 20.90$^{*}$ & ---
& --- & --- \\

& RNNLogic
& 0.5625 & 49.56 & 61.80
& 0.3276 & 23.49 & 38.60
& 4.5 h & 0.6 h \\

\addlinespace[4pt]

\multirow{2}{*}{\shortstack[l]{Path-based\\reasoning}}
& PRA / PathRank
& 0.4678 & 37.39 & 55.22
& 0.0972 & 6.35 & 10.72
& 239 s & 229 s \\

& MINERVA
& --- & --- & ---
& 0.2734 & 19.77 & 30.34
& --- & 3.3 h \\

\addlinespace[4pt]

GNN-based
& NBFNet
& 0.4946 & 37.96 & 55.54
& \textbf{0.5114} & \textbf{41.64} & \textbf{55.95}
& 2.3 h & 10.4 h \\

\midrule

Brain-inspired
& NSR
& 0.5893 & 52.80 & 64.32
& 0.3649 & 28.45 & 39.74
& 0.3 h & 0.6 h \\

\bottomrule
\end{tabular}%
}

\vspace{2pt}
\parbox{\textwidth}{\footnotesize
$^{*}$Values reported in prior work or official releases.
The AnyBURL FB15k-237 results and its 1000 s rule-learning budget
come from the official results page
(\url{https://web.informatik.uni-mannheim.de/AnyBURL/});
NCRL values come from \cite{cheng2023neural}.
Published and cross-hardware timings are indicative rather than
strictly matched; The reported NSR training time correspond to the accelerated implementation for static knowledge graphs described in Appendix~\ref{app:offline-pseudo}.
}
\end{table*}

Across the four benchmarks, NSR delivers competitive predictive performance while maintaining strong training efficiency. It ranks near the top on Nations and YAGO3-10 and remains competitive on FB15k-237, although its performance is weaker on Kinship, indicating dataset-dependent strengths rather than uniformly superior accuracy. At the same time, NSR trains substantially faster than several neural baselines and remains efficient even on the larger graphs. Overall, these results suggest that NSR offers a favorable trade-off between reasoning performance and computational cost, making it a competitive structural reasoner for knowledge-graph tasks where both accuracy and training efficiency matter.

\subsection{Traceable reasoning steps}

\begin{figure}[htbp]
    \centering
    \includegraphics[width=1\linewidth]{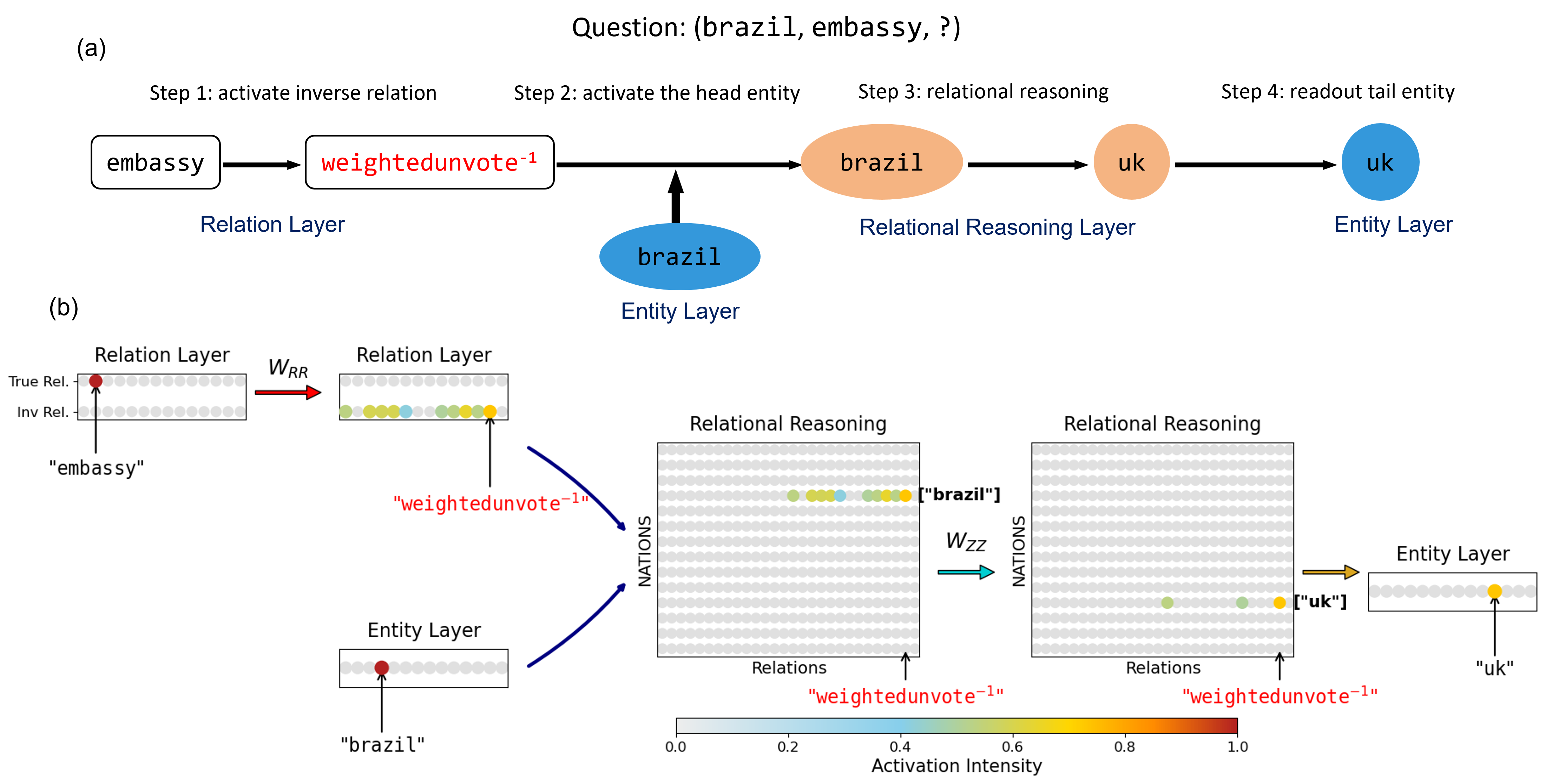}
    \caption{\textbf{Traceable reasoning in NSR.} 
 \textbf{(a)} A schematic of the inferred logical path: \texttt{embassy} triggers its inverse to map \texttt{Brazil} to \texttt{UK}. 
\textbf{(b)} Neuronal activations at each stage of the query \texttt{(Brazil, embassy, ?)}. Activating the relation neuron \texttt{embassy} triggers the activation of candidate equivalent relations (e.g., \texttt{weightedunvote}$^{-1}$ is an associated inverse relation of \texttt{embassy}) in the $L_R$ layer. 
The selected relation and head entity \texttt{Brazil} jointly project onto the $L_Z$ layer. State transition in $L_Z$ through recurrent dynamics leads to activation of neurons  encoding the tail entity \texttt{UK}. }
    \label{fig:TRstep}
\end{figure}


Unlike embedding baselines that rely on post-hoc attention or gradient saliency to explain opaque scalar scores, NSR's interpretability is intrinsic: the activation trajectory \emph{is} the computation. For the query \texttt{(Brazil, embassy, ?)} (Figure~\ref{fig:TRstep}), Step~1 of \S\ref{sec:reasoning-phase} clamps \texttt{embassy} onto the $L_R$ layer and iterates the $L_R$ dynamics once through $W_{RR}$. The Hebbian couplings learned during training (Eq.~\eqref{eq:hebb}) activate \texttt{weightedunvote}$^{-1}$ as the dominant equivalent relation, and its activation level serves as the confidence score $S_R$. Step~2 then activates the head entity \texttt{Brazil} in $L_E$ while holding \texttt{weightedunvote}$^{-1}$ active in $L_R$; the joint input projects onto the $L_Z$ relation map via the $W_{E\leftrightarrow Z}$ and $W_{Z\leftrightarrow R}$, forming the initial state $\boldsymbol{X}_Z^{\text{init}}$. Recurrent dynamics through $W_{ZZ}$ concentrate activation onto the neuron $z_{\text{UK}}^{\text{weightedunvote}^{-1}}$, which is read out in $L_E$ as the tail entity \texttt{UK}. Because each active unit represents a discrete, human-readable proposition, errors are auditable by inspecting which relation was activated in $L_R$ or where the $L_Z$ dynamics diverged. The resulting activation sequence provides an inspectable computational trace, inspired by evidence of non-spatial task-state replay in the human hippocampus~\cite{schuck2019sequential}. Symmetric trace is given in Appendix~\ref{app:additional-traces}.

\subsection{Learning of latent structures}

\begin{figure}[htbp]
    \centering
    \includegraphics[width=0.9\linewidth]{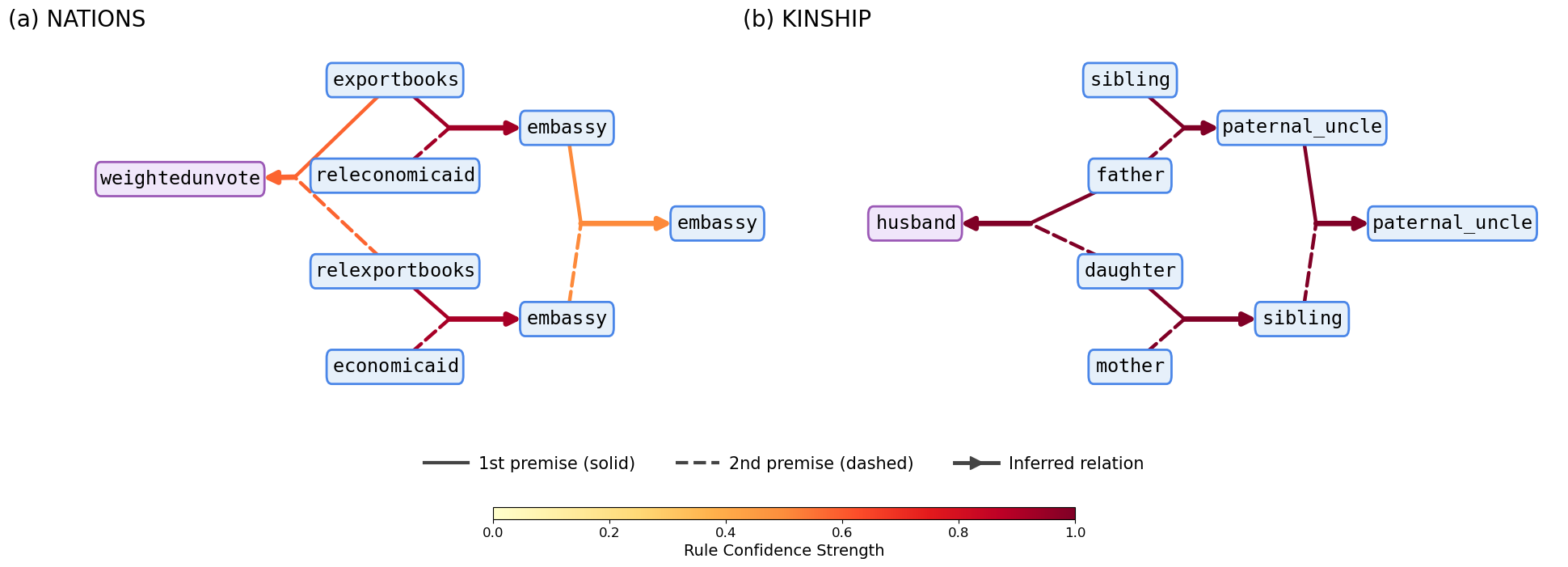}
    \caption{\textbf{Latent relational hierarchy.} 
    Compositional rules learned by NSR, satisfying $(h,r_1,x)\wedge(x,r_2,t) = (h,r_3,t)$, on (a) Nations and (b) Kinship1990\_EXTENDED. 
    Solid lines denote $r_1$, dashed lines denote $r_2$, and color intensity indicates rule confidence. }
    \label{fig:relations_learning}
\end{figure}

\paragraph{Latent relational hierarchy.}  
NSR recovers the latent compositional structure of the relational space rather than memorizing pairwise co-occurrences. As shown in Figure~\ref{fig:relations_learning}, the model learns which relation sequences form reliable multi-hop pathways. For example, the chain \texttt{exportbooks} $+$ \texttt{releconomicaid}  $\to$ \texttt{embassy} attains confidence 0.93 because the intermediate transition constitutes a coherent diplomatic pathway, whereas merely repeating \texttt{embassy} twice fails to yield a high-confidence rule despite the relation's high frequency. This selectivity shows that composition is context-sensitive and depends on whether a relation pair creates a semantically stable bridge. Notably, on Kinship1990\_EXTENDED the model can assign confidence 1 to deterministic kinship rules after a single observation by integrating prior logical constraints through hyperparameters, demonstrating that NSR is not limited to statistical induction.

\begin{figure}[htbp]
    \centering
    \includegraphics[width=0.65\linewidth]{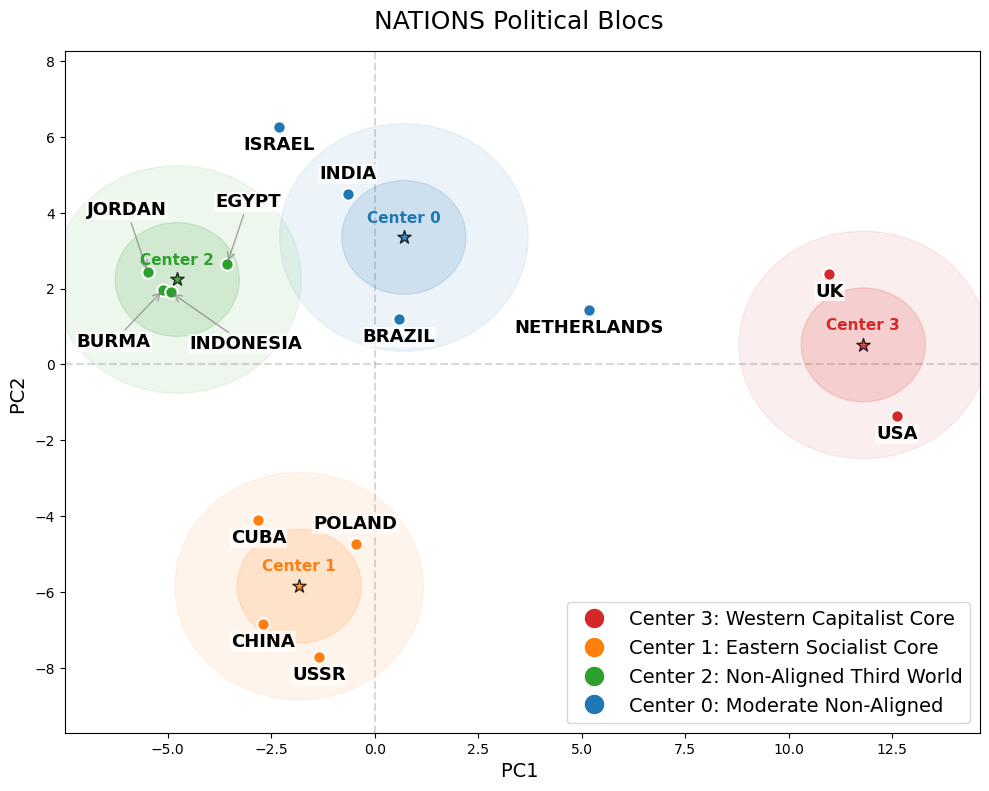}
    \caption{\textbf{Learned relational organization in Nations.} 
    PCA projection of nations based on their relational connections in $L_Z$. Four clusters emerge corresponding to Cold War political blocs.}
    \label{fig:Nations_Clusters}
\end{figure}

\paragraph{Relational organization of entities.}  
NSR encodes entities not as isolated symbols but as nodes characterized by their recurrent connections $L_{Z}$. As illustrated in Figure~\ref{fig:Nations_Clusters}, projecting entities based on their relational connections ($W_{ZZ}$) induces structurally meaningful clusters: the Western capitalist core (UK, USA), the Eastern socialist core (USSR, China), and distinct non-aligned blocs (Jordan, Egypt, India, Brazil) each occupy separate regions of the latent space. Taken together, NSR learns a dual latent structure: a compositional hierarchy governing how relations combine, and a relational topology governing how entities are organized.

\section{Conclusion}

We presented Neural Structural Reasoner (NSR), a brain-inspired architecture that integrates hierarchical representation with relation-driven state transition and Hebbian-style learning to accomplish structural reasoning over knowledge graphs. By preserving relational structure in layered network connectivity rather than collapsing it into flat embeddings, NSR achieves competitive, dataset-dependent accuracy on standard benchmarks, while its inference dynamics natively expose human-readable reasoning traces. NSR also spontaneously discovers interpretable latent structures such as compositional rules among relations and relational similarity among entities. These properties make NSR a promising step towards building efficient and interpretable models for structural reasoning. 

NSR is currently evaluated on discrete triplet-structured knowledge graphs, which provide a controlled setting for structural reasoning but leave open whether the same mechanisms generalize to temporal, hyper-relational, event-based, or noisy real-world knowledge. The model also assumes that entities and relations are already symbolically specified, so future work should couple NSR with neural modules that extract relational structure from text, perception, or episodic experience. While NSR is efficient on current benchmarks, scaling compositional-rule discovery to larger graphs and longer multi-hop chains will require sparse activation, approximate retrieval, or learned proposal mechanisms. Future work should also test robustness to missing or contradictory triples and develop more biologically grounded implementations of sequence detection, gating, and confidence aggregation.

\section*{Acknowledgments and Disclosures}

We thank Zexi Su, Hehao Qin for insightful discussions and helpful feedback on this work.

\textbf{Funding.}
This work was supported by the National Science and Technology Major Project on Brain Science and Brain-Like Intelligence Technology (Project No.~2025ZD0217400), the CAMS Innovation Fund for Medical Sciences (CIFMS; No.~2024-RC180-02), the Beijing Key Laboratory of Brain Science and Brain-Machine Interface, and the Fundamental and Interdisciplinary Disciplines Breakthrough Plan of the Ministry of Education of China (No.~JYB2025XDXM504). 

\textbf{Competing Interests.}
The authors declare no competing interests.
\clearpage
\bibliographystyle{plainnat}
\bibliography{ref}

\begin{thebibliography}{62}
\providecommand{\natexlab}[1]{#1}
\providecommand{\url}[1]{\texttt{#1}}
\expandafter\ifx\csname urlstyle\endcsname\relax
  \providecommand{\doi}[1]{doi: #1}\else
  \providecommand{\doi}{doi: \begingroup \urlstyle{rm}\Url}\fi

\bibitem[Akiba et~al.(2019)Akiba, Sano, Yanase, Ohta, and Koyama]{akiba2019optuna}
Takuya Akiba, Shotaro Sano, Toshihiko Yanase, Takeru Ohta, and Masanori Koyama.
\newblock Optuna: A next-generation hyperparameter optimization framework.
\newblock In \emph{Proceedings of the 25th ACM SIGKDD international conference on knowledge discovery \& data mining}, pages 2623--2631, 2019.

\bibitem[Ali et~al.(2021)Ali, Berrendorf, Hoyt, Vermue, Sharifzadeh, Tresp, and Lehmann]{ali2021pykeen}
Mehdi Ali, Max Berrendorf, Charles~Tapley Hoyt, Laurent Vermue, Sahand Sharifzadeh, Volker Tresp, and Jens Lehmann.
\newblock Pykeen 1.0: a python library for training and evaluating knowledge graph embeddings.
\newblock \emph{Journal of Machine Learning Research}, 22\penalty0 (82):\penalty0 1--6, 2021.

\bibitem[Barrett et~al.(2018)Barrett, Hill, Santoro, Morcos, and Lillicrap]{barrett2018}
David Barrett, Felix Hill, Adam Santoro, Ari Morcos, and Timothy Lillicrap.
\newblock Measuring abstract reasoning in neural networks.
\newblock In Jennifer Dy and Andreas Krause, editors, \emph{Proceedings of the 35th International Conference on Machine Learning}, volume~80 of \emph{Proceedings of Machine Learning Research}, pages 511--520. PMLR, 10--15 Jul 2018.
\newblock URL \url{https://proceedings.mlr.press/v80/barrett18a.html}.

\bibitem[Battaglia et~al.(2018)Battaglia, Hamrick, Bapst, Sanchez-Gonzalez, Zambaldi, Malinowski, Tacchetti, Raposo, Santoro, Faulkner, Gulcehre, Song, Ballard, Gilmer, Dahl, Vaswani, Allen, Nash, Langston, Dyer, Heess, Wierstra, Kohli, Botvinick, Vinyals, Li, and Pascanu]{battaglia2018}
Peter~W. Battaglia, Jessica~B. Hamrick, Victor Bapst, Alvaro Sanchez-Gonzalez, Vinicius Zambaldi, Mateusz Malinowski, Andrea Tacchetti, David Raposo, Adam Santoro, Ryan Faulkner, Caglar Gulcehre, Francis Song, Andrew Ballard, Justin Gilmer, George Dahl, Ashish Vaswani, Kelsey Allen, Charles Nash, Victoria Langston, Chris Dyer, Nicolas Heess, Daan Wierstra, Pushmeet Kohli, Matt Botvinick, Oriol Vinyals, Yujia Li, and Razvan Pascanu.
\newblock Relational inductive biases, deep learning, and graph networks, 2018.
\newblock URL \url{https://arxiv.org/abs/1806.01261}.

\bibitem[Behrens et~al.(2018)Behrens, Muller, Whittington, Mark, Baram, Stachenfeld, and Kurth-Nelson]{behrens2018}
Timothy~E.J. Behrens, Timothy~H. Muller, James~C.R. Whittington, Shirley Mark, Alon~B. Baram, Kimberly~L. Stachenfeld, and Zeb Kurth-Nelson.
\newblock What is a cognitive map? organizing knowledge for flexible behavior.
\newblock \emph{Neuron}, 100\penalty0 (2):\penalty0 490--509, 2018.
\newblock ISSN 0896-6273.
\newblock \doi{https://doi.org/10.1016/j.neuron.2018.10.002}.
\newblock URL \url{https://www.sciencedirect.com/science/article/pii/S0896627318308560}.

\bibitem[Bordes et~al.(2013)Bordes, Usunier, Garcia-Duran, Weston, and Yakhnenko]{bordes2013translating}
Antoine Bordes, Nicolas Usunier, Alberto Garcia-Duran, Jason Weston, and Oksana Yakhnenko.
\newblock Translating embeddings for modeling multi-relational data.
\newblock \emph{Advances in neural information processing systems}, 26, 2013.

\bibitem[Brachman and Levesque(2004)]{brachman2004knowledge}
Ronald Brachman and Hector Levesque.
\newblock \emph{Knowledge representation and reasoning}.
\newblock Elsevier, 2004.

\bibitem[Bunge et~al.(2005)Bunge, Wendelken, Badre, and Wagner]{bunge2005analogical}
Silvia~A Bunge, Carter Wendelken, David Badre, and Anthony~D Wagner.
\newblock Analogical reasoning and prefrontal cortex: evidence for separable retrieval and integration mechanisms.
\newblock \emph{Cerebral cortex}, 15\penalty0 (3):\penalty0 239--249, 2005.

\bibitem[Burak and Fiete(2009)]{burak2009accurate}
Yoram Burak and Ila~R Fiete.
\newblock Accurate path integration in continuous attractor network models of grid cells.
\newblock \emph{PLoS computational biology}, 5\penalty0 (2):\penalty0 e1000291, 2009.

\bibitem[Chandra et~al.(2025)Chandra, Sharma, Chaudhuri, and Fiete]{chandra2025episodic}
Sarthak Chandra, Sugandha Sharma, Rishidev Chaudhuri, and Ila Fiete.
\newblock Episodic and associative memory from spatial scaffolds in the hippocampus.
\newblock \emph{Nature}, 638\penalty0 (8051):\penalty0 739--751, 2025.

\bibitem[Cheng et~al.(2023)Cheng, Ahmed, and Sun]{cheng2023neural}
Kewei Cheng, Nesreen~K Ahmed, and Yizhou Sun.
\newblock Neural compositional rule learning for knowledge graph reasoning.
\newblock \emph{arXiv preprint arXiv:2303.03581}, 2023.

\bibitem[Christoff et~al.(2001)Christoff, Prabhakaran, Dorfman, Zhao, Kroger, Holyoak, and Gabrieli]{christoff2001rostrolateral}
Kalina Christoff, Vivek Prabhakaran, Jennifer Dorfman, Zuo Zhao, James~K Kroger, Keith~J Holyoak, and John~DE Gabrieli.
\newblock Rostrolateral prefrontal cortex involvement in relational integration during reasoning.
\newblock \emph{Neuroimage}, 14\penalty0 (5):\penalty0 1136--1149, 2001.

\bibitem[Constantinescu et~al.(2016)Constantinescu, O’Reilly, and Behrens]{constantinescu2016organizing}
Alexandra~O Constantinescu, Jill~X O’Reilly, and Timothy~EJ Behrens.
\newblock Organizing conceptual knowledge in humans with a gridlike code.
\newblock \emph{Science}, 352\penalty0 (6292):\penalty0 1464--1468, 2016.

\bibitem[Das et~al.(2017)Das, Dhuliawala, Zaheer, Vilnis, Durugkar, Krishnamurthy, Smola, and McCallum]{das2017go}
Rajarshi Das, Shehzaad Dhuliawala, Manzil Zaheer, Luke Vilnis, Ishan Durugkar, Akshay Krishnamurthy, Alex Smola, and Andrew McCallum.
\newblock Go for a walk and arrive at the answer: Reasoning over paths in knowledge bases using reinforcement learning.
\newblock \emph{arXiv preprint arXiv:1711.05851}, 2017.

\bibitem[Dettmers et~al.(2018)Dettmers, Minervini, Stenetorp, and Riedel]{dettmers2018convolutional}
Tim Dettmers, Pasquale Minervini, Pontus Stenetorp, and Sebastian Riedel.
\newblock Convolutional 2d knowledge graph embeddings.
\newblock In \emph{Proceedings of the AAAI conference on artificial intelligence}, volume~32, 2018.

\bibitem[Dusek and Eichenbaum(1997)]{dusek1997hippocampus}
Jeffery~A Dusek and Howard Eichenbaum.
\newblock The hippocampus and memory for orderly stimulus relations.
\newblock \emph{Proceedings of the National Academy of Sciences}, 94\penalty0 (13):\penalty0 7109--7114, 1997.

\bibitem[Dziri et~al.(2023)Dziri, Lu, Sclar, Li, Jiang, Lin, Welleck, West, Bhagavatula, Le~Bras, Hwang, Sanyal, Ren, Ettinger, Harchaoui, and Choi]{dziri2023}
Nouha Dziri, Ximing Lu, Melanie Sclar, Xiang~(Lorraine) Li, Liwei Jiang, Bill~Yuchen Lin, Sean Welleck, Peter West, Chandra Bhagavatula, Ronan Le~Bras, Jena Hwang, Soumya Sanyal, Xiang Ren, Allyson Ettinger, Zaid Harchaoui, and Yejin Choi.
\newblock Faith and fate: Limits of transformers on compositionality.
\newblock In A.~Oh, T.~Naumann, A.~Globerson, K.~Saenko, M.~Hardt, and S.~Levine, editors, \emph{Advances in Neural Information Processing Systems}, volume~36, pages 70293--70332. Curran Associates, Inc., 2023.
\newblock URL \url{https://proceedings.neurips.cc/paper_files/paper/2023/file/deb3c28192f979302c157cb653c15e90-Paper-Conference.pdf}.

\bibitem[Eichenbaum and Cohen(2004)]{eichenbaum2004conditioning}
Howard Eichenbaum and Neal~J Cohen.
\newblock \emph{From conditioning to conscious recollection: Memory systems of the brain}.
\newblock Number~35. Oxford university press, 2004.

\bibitem[Fodor and Pylyshyn(1988)]{fodor1988}
Jerry~A. Fodor and Zenon~W. Pylyshyn.
\newblock Connectionism and cognitive architecture: A critical analysis.
\newblock \emph{Cognition}, 28\penalty0 (1):\penalty0 3--71, 1988.
\newblock ISSN 0010-0277.
\newblock \doi{https://doi.org/10.1016/0010-0277(88)90031-5}.
\newblock URL \url{https://www.sciencedirect.com/science/article/pii/0010027788900315}.

\bibitem[Gal{\'a}rraga et~al.(2013)Gal{\'a}rraga, Teflioudi, Hose, and Suchanek]{galarraga2013amie}
Luis~Antonio Gal{\'a}rraga, Christina Teflioudi, Katja Hose, and Fabian Suchanek.
\newblock Amie: association rule mining under incomplete evidence in ontological knowledge bases.
\newblock In \emph{Proceedings of the 22nd international conference on World Wide Web}, pages 413--422, 2013.

\bibitem[Galkin et~al.(2023)Galkin, Yuan, Mostafa, Tang, and Zhu]{galkin2023towards}
Mikhail Galkin, Xinyu Yuan, Hesham Mostafa, Jian Tang, and Zhaocheng Zhu.
\newblock Towards foundation models for knowledge graph reasoning.
\newblock \emph{arXiv preprint arXiv:2310.04562}, 2023.

\bibitem[Gentner(1983)]{gentner1983}
Dedre Gentner.
\newblock Structure-mapping: A theoretical framework for analogy.
\newblock \emph{Cognitive Science}, 7\penalty0 (2):\penalty0 155--170, 1983.
\newblock ISSN 0364-0213.
\newblock \doi{https://doi.org/10.1016/S0364-0213(83)80009-3}.
\newblock URL \url{https://www.sciencedirect.com/science/article/pii/S0364021383800093}.

\bibitem[He et~al.(2026)He, Li, White, and Vitercik]{he2026}
Yu~He, Yingxi Li, Colin White, and Ellen Vitercik.
\newblock Can llms reason structurally? benchmarking via the lens of data structures, 2026.
\newblock URL \url{https://arxiv.org/abs/2505.24069}.

\bibitem[Hogan et~al.(2021)Hogan, Blomqvist, Cochez, D’amato, Melo, Gutierrez, Kirrane, Gayo, Navigli, Neumaier, Ngomo, Polleres, Rashid, Rula, Schmelzeisen, Sequeda, Staab, and Zimmermann]{hogan2021}
Aidan Hogan, Eva Blomqvist, Michael Cochez, Claudia D’amato, Gerard~De Melo, Claudio Gutierrez, Sabrina Kirrane, Jos\'{e} Emilio~Labra Gayo, Roberto Navigli, Sebastian Neumaier, Axel-Cyrille~Ngonga Ngomo, Axel Polleres, Sabbir~M. Rashid, Anisa Rula, Lukas Schmelzeisen, Juan Sequeda, Steffen Staab, and Antoine Zimmermann.
\newblock Knowledge graphs.
\newblock \emph{ACM Comput. Surv.}, 54\penalty0 (4), July 2021.
\newblock ISSN 0360-0300.
\newblock \doi{10.1145/3447772}.
\newblock URL \url{https://doi.org/10.1145/3447772}.

\bibitem[Ji et~al.(2022)Ji, Pan, Cambria, Marttinen, and Yu]{ji2022}
Shaoxiong Ji, Shirui Pan, Erik Cambria, Pekka Marttinen, and Philip~S. Yu.
\newblock A survey on knowledge graphs: Representation, acquisition, and applications.
\newblock \emph{IEEE Transactions on Neural Networks and Learning Systems}, 33\penalty0 (2):\penalty0 494--514, 2022.
\newblock \doi{10.1109/TNNLS.2021.3070843}.

\bibitem[Kemp et~al.(2006)Kemp, Tenenbaum, Griffiths, Yamada, and Ueda]{kemp2006learning}
Charles Kemp, Joshua~B Tenenbaum, Thomas~L Griffiths, Takeshi Yamada, and Naonori Ueda.
\newblock Learning systems of concepts with an infinite relational model.
\newblock In \emph{AAAI}, volume~3, page~5, 2006.

\bibitem[Lake et~al.(2017)Lake, Ullman, Tenenbaum, and Gershman]{brenden2017}
Brenden~M. Lake, Tomer~D. Ullman, Joshua~B. Tenenbaum, and Samuel~J. Gershman.
\newblock Building machines that learn and think like people.
\newblock \emph{Behavioral and Brain Sciences}, 40:\penalty0 e253, 2017.
\newblock \doi{10.1017/S0140525X16001837}.

\bibitem[Lao et~al.(2011)Lao, Mitchell, and Cohen]{lao2011random}
Ni~Lao, Tom Mitchell, and William Cohen.
\newblock Random walk inference and learning in a large scale knowledge base.
\newblock In \emph{Proceedings of the 2011 conference on empirical methods in natural language processing}, pages 529--539, 2011.

\bibitem[Meilicke et~al.(2024)Meilicke, Chekol, Betz, Fink, and Stuckenschmidt]{meilicke2024anytime}
Christian Meilicke, Melisachew~Wudage Chekol, Patrick Betz, Manuel Fink, and Heiner Stuckenschmidt.
\newblock Anytime bottom-up rule learning for large-scale knowledge graph completion: C. meilicke et al.
\newblock \emph{The VLDB Journal}, 33\penalty0 (1):\penalty0 131--161, 2024.

\bibitem[Nickel et~al.(2011)Nickel, Tresp, Kriegel, et~al.]{nickel2011three}
Maximilian Nickel, Volker Tresp, Hans-Peter Kriegel, et~al.
\newblock A three-way model for collective learning on multi-relational data.
\newblock In \emph{Icml}, volume~11, pages 3104482--3104584, 2011.

\bibitem[Nickel et~al.(2016)Nickel, Rosasco, and Poggio]{nickel2016holographic}
Maximilian Nickel, Lorenzo Rosasco, and Tomaso Poggio.
\newblock Holographic embeddings of knowledge graphs.
\newblock In \emph{Proceedings of the AAAI conference on artificial intelligence}, volume~30, 2016.

\bibitem[Niu(2024)]{niu2024knowledge}
Guanglin Niu.
\newblock Knowledge graph embeddings: A comprehensive survey on capturing relation properties.
\newblock \emph{arXiv preprint arXiv:2410.14733}, 2024.

\bibitem[Paccanaro and Hinton(2001)]{917563}
A.~Paccanaro and G.E. Hinton.
\newblock Learning distributed representations of concepts using linear relational embedding.
\newblock \emph{IEEE Transactions on Knowledge and Data Engineering}, 13\penalty0 (2):\penalty0 232--244, 2001.
\newblock \doi{10.1109/69.917563}.

\bibitem[Pan et~al.(2025)Pan, D’Ambrogio, Kingston, Rascu, Sankhe, Luo, Klein-Fl{\"u}gge, Mahmoodi, and Rushworth]{pan2025causal}
Deng Pan, Simone D’Ambrogio, Naomi Kingston, Miruna Rascu, Pranav Sankhe, Shuyi Luo, Miriam~C Klein-Fl{\"u}gge, Ali Mahmoodi, and Matthew~FS Rushworth.
\newblock Causal necessity of human hippocampus for structure-based inference in learning.
\newblock \emph{bioRxiv}, pages 2025--08, 2025.

\bibitem[Park et~al.(2021)Park, Miller, and Boorman]{park2021inferences}
Seongmin~A Park, Douglas~S Miller, and Erie~D Boorman.
\newblock Inferences on a multidimensional social hierarchy use a grid-like code.
\newblock \emph{Nature neuroscience}, 24\penalty0 (9):\penalty0 1292--1301, 2021.

\bibitem[Patterson et~al.(2007)Patterson, Nestor, and Rogers]{patterson2007you}
Karalyn Patterson, Peter~J Nestor, and Timothy~T Rogers.
\newblock Where do you know what you know? the representation of semantic knowledge in the human brain.
\newblock \emph{Nature reviews neuroscience}, 8\penalty0 (12):\penalty0 976--987, 2007.

\bibitem[Press et~al.(2023)Press, Zhang, Min, Schmidt, Smith, and Lewis]{press2023}
Ofir Press, Muru Zhang, Sewon Min, Ludwig Schmidt, Noah Smith, and Mike Lewis.
\newblock Measuring and narrowing the compositionality gap in language models.
\newblock In Houda Bouamor, Juan Pino, and Kalika Bali, editors, \emph{Findings of the Association for Computational Linguistics: EMNLP 2023}, pages 5687--5711, Singapore, December 2023. Association for Computational Linguistics.
\newblock \doi{10.18653/v1/2023.findings-emnlp.378}.
\newblock URL \url{https://aclanthology.org/2023.findings-emnlp.378/}.

\bibitem[Preston et~al.(2004)Preston, Shrager, Dudukovic, and Gabrieli]{preston2004hippocampal}
Alison~R Preston, Yael Shrager, Nicole~M Dudukovic, and John~DE Gabrieli.
\newblock Hippocampal contribution to the novel use of relational information in declarative memory.
\newblock \emph{Hippocampus}, 2004.

\bibitem[Qu et~al.(2021)Qu, Chen, Xhonneux, Bengio, and Tang]{qu2021rnnlogic}
Meng Qu, Junkun Chen, Louis-Pascal Xhonneux, Yoshua Bengio, and Jian Tang.
\newblock {\{}RNNL{\}}ogic: Learning logic rules for reasoning on knowledge graphs.
\newblock In \emph{International Conference on Learning Representations}, 2021.
\newblock URL \url{https://openreview.net/forum?id=tGZu6DlbreV}.

\bibitem[Ralph et~al.(2017)Ralph, Jefferies, Patterson, and Rogers]{ralph2017neural}
Matthew A~Lambon Ralph, Elizabeth Jefferies, Karalyn Patterson, and Timothy~T Rogers.
\newblock The neural and computational bases of semantic cognition.
\newblock \emph{Nature reviews neuroscience}, 18\penalty0 (1):\penalty0 42--55, 2017.

\bibitem[Ramsauer et~al.(2021)Ramsauer, Sch{\"a}fl, Lehner, Seidl, Widrich, Gruber, Holzleitner, Adler, Kreil, Kopp, Klambauer, Brandstetter, and Hochreiter]{ramsauer2021hopfield}
Hubert Ramsauer, Bernhard Sch{\"a}fl, Johannes Lehner, Philipp Seidl, Michael Widrich, Lukas Gruber, Markus Holzleitner, Thomas Adler, David Kreil, Michael~K Kopp, G{\"u}nter Klambauer, Johannes Brandstetter, and Sepp Hochreiter.
\newblock Hopfield networks is all you need.
\newblock In \emph{International Conference on Learning Representations}, 2021.
\newblock URL \url{https://openreview.net/forum?id=tL89RnzIiCd}.

\bibitem[Samborska et~al.(2022)Samborska, Butler, Walton, Behrens, and Akam]{samborska2022complementary}
Veronika Samborska, James~L Butler, Mark~E Walton, Timothy~EJ Behrens, and Thomas Akam.
\newblock Complementary task representations in hippocampus and prefrontal cortex for generalizing the structure of problems.
\newblock \emph{Nature Neuroscience}, 25\penalty0 (10):\penalty0 1314--1326, 2022.

\bibitem[Santoro et~al.(2017)Santoro, Raposo, Barrett, Malinowski, Pascanu, Battaglia, and Lillicrap]{santoro2017}
Adam Santoro, David Raposo, David~G Barrett, Mateusz Malinowski, Razvan Pascanu, Peter Battaglia, and Timothy Lillicrap.
\newblock A simple neural network module for relational reasoning.
\newblock In I.~Guyon, U.~Von Luxburg, S.~Bengio, H.~Wallach, R.~Fergus, S.~Vishwanathan, and R.~Garnett, editors, \emph{Advances in Neural Information Processing Systems}, volume~30. Curran Associates, Inc., 2017.
\newblock URL \url{https://proceedings.neurips.cc/paper_files/paper/2017/file/e6acf4b0f69f6f6e60e9a815938aa1ff-Paper.pdf}.

\bibitem[Schlichtkrull et~al.(2018)Schlichtkrull, Kipf, Bloem, Van Den~Berg, Titov, and Welling]{schlichtkrull2018modeling}
Michael Schlichtkrull, Thomas~N Kipf, Peter Bloem, Rianne Van Den~Berg, Ivan Titov, and Max Welling.
\newblock Modeling relational data with graph convolutional networks.
\newblock In \emph{European semantic web conference}, pages 593--607. Springer, 2018.

\bibitem[Schuck and Niv(2019)]{schuck2019sequential}
Nicolas~W Schuck and Yael Niv.
\newblock Sequential replay of nonspatial task states in the human hippocampus.
\newblock \emph{Science}, 364\penalty0 (6447):\penalty0 eaaw5181, 2019.

\bibitem[Stachenfeld et~al.(2017)Stachenfeld, Botvinick, and Gershman]{stachenfeld2017}
Kimberly~L Stachenfeld, Matthew~M Botvinick, and Samuel~J Gershman.
\newblock The hippocampus as a predictive map.
\newblock \emph{Nature neuroscience}, 20\penalty0 (11):\penalty0 1643--1653, 2017.

\bibitem[Steinberg and Sompolinsky(2022)]{steinberg2022associative}
Julia Steinberg and Haim Sompolinsky.
\newblock Associative memory of structured knowledge.
\newblock \emph{Scientific Reports}, 12\penalty0 (1):\penalty0 21808, 2022.

\bibitem[Sun et~al.(2019)Sun, Deng, Nie, and Tang]{sun2018rotate}
Zhiqing Sun, Zhi-Hong Deng, Jian-Yun Nie, and Jian Tang.
\newblock Rotate: Knowledge graph embedding by relational rotation in complex space.
\newblock In \emph{International Conference on Learning Representations}, 2019.
\newblock URL \url{https://openreview.net/forum?id=HkgEQnRqYQ}.

\bibitem[Tahir~Chaudhry et~al.(2024)Tahir~Chaudhry, Zavatone-Veth, Krotov, and Pehlevan]{tahir2024long}
Hamza Tahir~Chaudhry, Jacob~A Zavatone-Veth, Dmitry Krotov, and Cengiz Pehlevan.
\newblock Long sequence hopfield memory.
\newblock \emph{Journal of Statistical Mechanics: Theory and Experiment}, 2024\penalty0 (10):\penalty0 104024, 2024.

\bibitem[Toutanova and Chen(2015)]{toutanova2015observed}
Kristina Toutanova and Danqi Chen.
\newblock Observed versus latent features for knowledge base and text inference.
\newblock In \emph{Proceedings of the 3rd workshop on continuous vector space models and their compositionality}, pages 57--66, 2015.

\bibitem[Trouillon et~al.(2016)Trouillon, Welbl, Riedel, Gaussier, and Bouchard]{trouillon2016complex}
Th{\'e}o Trouillon, Johannes Welbl, Sebastian Riedel, {\'E}ric Gaussier, and Guillaume Bouchard.
\newblock Complex embeddings for simple link prediction.
\newblock In \emph{International conference on machine learning}, pages 2071--2080. PMLR, 2016.

\bibitem[Velivckovi'c et~al.(2022)Velivckovi'c, Badia, Budden, Pascanu, Banino, Dashevskiy, Hadsell, and Blundell]{petar2022}
Petar Velivckovi'c, Adri{\`a}~Puigdom{\`e}nech Badia, David Budden, Razvan Pascanu, Andrea Banino, Mikhail Dashevskiy, Raia Hadsell, and Charles Blundell.
\newblock The clrs algorithmic reasoning benchmark.
\newblock In \emph{International Conference on Machine Learning}, 2022.
\newblock URL \url{https://api.semanticscholar.org/CorpusID:249210177}.

\bibitem[Waltz et~al.(1999)Waltz, Knowlton, Holyoak, Boone, Mishkin, de~Menezes~Santos, Thomas, and Miller]{waltz1999system}
James~A Waltz, Barbara~J Knowlton, Keith~J Holyoak, Kyle~B Boone, Fred~S Mishkin, Marcia de~Menezes~Santos, Carmen~R Thomas, and Bruce~L Miller.
\newblock A system for relational reasoning in human prefrontal cortex.
\newblock \emph{Psychological science}, 10\penalty0 (2):\penalty0 119--125, 1999.

\bibitem[Wang et~al.(2017)Wang, Mao, Wang, and Guo]{wang2017}
Quan Wang, Zhendong Mao, Bin Wang, and Li~Guo.
\newblock Knowledge graph embedding: A survey of approaches and applications.
\newblock \emph{IEEE Transactions on Knowledge and Data Engineering}, 29\penalty0 (12):\penalty0 2724--2743, 2017.
\newblock \doi{10.1109/TKDE.2017.2754499}.

\bibitem[Webb et~al.(2024)Webb, Frankland, Altabaa, Segert, Krishnamurthy, Campbell, Russin, Giallanza, O’Reilly, Lafferty, and Cohen]{webb2024}
Taylor~W. Webb, Steven~M. Frankland, Awni Altabaa, Simon Segert, Kamesh Krishnamurthy, Declan Campbell, Jacob Russin, Tyler Giallanza, Randall O’Reilly, John Lafferty, and Jonathan~D. Cohen.
\newblock The relational bottleneck as an inductive bias for efficient abstraction.
\newblock \emph{Trends in Cognitive Sciences}, 28\penalty0 (9):\penalty0 829--843, 2024.
\newblock ISSN 1364-6613.
\newblock \doi{https://doi.org/10.1016/j.tics.2024.04.001}.
\newblock URL \url{https://www.sciencedirect.com/science/article/pii/S1364661324000809}.

\bibitem[Wendelken et~al.(2008)Wendelken, Nakhabenko, Donohue, Carter, and Bunge]{wendelken2008brain}
Carter Wendelken, Denis Nakhabenko, Sarah~E Donohue, Cameron~S Carter, and Silvia~A Bunge.
\newblock “brain is to thought as stomach is to??”: investigating the role of rostrolateral prefrontal cortex in relational reasoning.
\newblock \emph{Journal of cognitive neuroscience}, 20\penalty0 (4):\penalty0 682--693, 2008.

\bibitem[Whittington et~al.(2020)Whittington, Muller, Mark, Chen, Barry, Burgess, and Behrens]{whittington2020}
James~C.R. Whittington, Timothy~H. Muller, Shirley Mark, Guifen Chen, Caswell Barry, Neil Burgess, and Timothy~E.J. Behrens.
\newblock The tolman-eichenbaum machine: Unifying space and relational memory through generalization in the hippocampal formation.
\newblock \emph{Cell}, 183\penalty0 (5):\penalty0 1249--1263.e23, 2020.
\newblock ISSN 0092-8674.
\newblock \doi{https://doi.org/10.1016/j.cell.2020.10.024}.
\newblock URL \url{https://www.sciencedirect.com/science/article/pii/S009286742031388X}.

\bibitem[Whittington et~al.(2021)Whittington, Warren, and Behrens]{whittington2021relating}
James~CR Whittington, Joseph Warren, and Timothy~EJ Behrens.
\newblock Relating transformers to models and neural representations of the hippocampal formation.
\newblock \emph{arXiv preprint arXiv:2112.04035}, 2021.

\bibitem[Wu et~al.(2023)Wu, Wan, Chen, Wu, Shen, and Lin]{wu2023}
Shuhan Wu, Huaiyu Wan, Wei Chen, Yuting Wu, Junfeng Shen, and Youfang Lin.
\newblock Towards enhancing relational rules for knowledge graph link prediction.
\newblock In Houda Bouamor, Juan Pino, and Kalika Bali, editors, \emph{Findings of the Association for Computational Linguistics: EMNLP 2023}, pages 10082--10097, Singapore, December 2023. Association for Computational Linguistics.
\newblock \doi{10.18653/v1/2023.findings-emnlp.676}.
\newblock URL \url{https://aclanthology.org/2023.findings-emnlp.676/}.

\bibitem[Yang et~al.(2017)Yang, Yang, and Cohen]{yang2017differentiable}
Fan Yang, Zhilin Yang, and William~W Cohen.
\newblock Differentiable learning of logical rules for knowledge base reasoning.
\newblock \emph{Advances in neural information processing systems}, 30, 2017.

\bibitem[Zhang et~al.(2020)Zhang, Yu, Saebi, Jiang, and Chawla]{zhang2020}
Chuxu Zhang, Lu~Yu, Mandana Saebi, Meng Jiang, and Nitesh Chawla.
\newblock Few-shot multi-hop relation reasoning over knowledge bases.
\newblock In Trevor Cohn, Yulan He, and Yang Liu, editors, \emph{Findings of the Association for Computational Linguistics: EMNLP 2020}, pages 580--585, Online, November 2020. Association for Computational Linguistics.
\newblock \doi{10.18653/v1/2020.findings-emnlp.51}.
\newblock URL \url{https://aclanthology.org/2020.findings-emnlp.51/}.

\bibitem[Zhu et~al.(2021)Zhu, Zhang, Xhonneux, and Tang]{zhu2021neural}
Zhaocheng Zhu, Zuobai Zhang, Louis-Pascal Xhonneux, and Jian Tang.
\newblock Neural bellman-ford networks: A general graph neural network framework for link prediction.
\newblock \emph{Advances in neural information processing systems}, 34:\penalty0 29476--29490, 2021.

\end{thebibliography}

\clearpage

\appendix

\appendix

\section{Details for Encoding and Learning}

\subsection{Learning and Encoding of KG Training Data}
\label{sec:learning-encoding}

This subsection details how NSR stores the training graph into its connectivity and extracts pairwise relational structure. Compositional multi-hop rules are deferred to \S\ref{sec:compositional-rules}.

\paragraph{Network initialization.} All plastic weights are initialized to zero. The only fixed connections are the bidirectional couplings between each entity neuron $e_i$ and its associated $L_Z$ neurons $z_i^k$ (Eq.~\eqref{eq:weightinit}).

\paragraph{Encoding and pairwise learning.} Each training triple $(h_i,r_k,t_j)$ and its inverse $(t_j,r_k^{-1},h_i)$ are encoded by setting the corresponding $L_Z$ intra-layer and $L_Z$--$L_R$ weights to one (Eq.~\eqref{eq:encw}). Concurrently, the network records which relations co-occur on the same $(h,t)$ pair; these co-activations drive the Oja-like updates of $W_{RR}$ (Eq.~\eqref{eq:hebb}) that capture equivalence, symmetry, and inverse structure.

\begin{algorithm}[htbp]
\caption{Learning and encoding of KG training data}
\label{alg:learning-encoding}
\small
\textbf{Input.} Training triples $\mathcal{T}_{\text{train}}$; $N$ entities; $M$ relations.\\
\textbf{Parameters.} Hebbian rate $\delta_{RR}$, decay $\mu_{RR}$.

\vspace{0.5em}
\noindent\begin{minipage}{\linewidth}
\raggedright
1: initialize all weight matrices to $0$\\
2: \textbf{for} $i=1,\dots,N$ \textbf{do}\\
3: \quad\textbf{for} $k=1,\dots,2M$ \textbf{do}\\
4: \quad\quad fix $W_{E\leftrightarrow Z}(e_i,z_i^k)=1$\\
5: \quad\textbf{end for}\\
6: \textbf{end for}\\
7: \textbf{for} each $(h_i,r_k,t_j)\in\mathcal{T}_{\text{train}}$ \textbf{do}\\
8: \quad set $W_{ZZ}(z_i^k,z_j^{k+M})=W_{ZZ}(z_j^{k+M},z_i^k)=1$\\
9: \quad set $W_{Z\leftrightarrow R}(z_i^k,r_k)=W_{Z\leftrightarrow R}(z_j^{k+M},r_{k+M})=1$\\
10: \quad record $r_k$ for pair $(h_i,t_j)$\\
11: \textbf{end for}\\
12: \textbf{for} each entity pair $(h,t)$ with $\ge 2$ recorded relations \textbf{do}\\
13: \quad \textbf{for} each distinct pair $(r_i,r_j)$ in that set \textbf{do}\\
14: \quad\quad $\Delta w_{ij}^{RR}\leftarrow\delta_{RR}(r_i r_j)-\mu_{RR}(r_i^2+r_j^2)w_{ij}^{RR}$\\
15: \quad\quad $w_{ij}^{RR}\leftarrow w_{ij}^{RR}+\Delta w_{ij}^{RR}$\\
16: \quad\textbf{end for}\\
17: \textbf{end for}
\end{minipage}
\end{algorithm}

\subsection{Compositional Rules Learning}
\label{sec:compositional-rules}

The Relation Composition layer $L_C$ contains sequence-selective neurons $\{c_{ij}\}$, each tuned to an ordered pair of extended relations $(\tilde r_i,\tilde r_j)$ with $\tilde r\in\{0,\dots,2M-1\}$ (forward $\tilde r_k\equiv r_k$ for $k<M$, inverse $\tilde r_{k+M}\equiv r_k^{-1}$). These neurons detect relational chains through asymmetric-delay coincidence detection.

\paragraph{Detection and encoding of relation composition.}
When the conditions in \eqref{eq:comp-rule} are satisfied, assign $L_C$ neurons $c_p$, $c_q$ and $c_k$ to encode relations $r_p$, $r_q$ and $r_k$ respectively. In addition, assign neuron $c_{\tilde{pq}}$ to be directly coupled with $c_q$ and indirectly coupled with $c_p$ through an intermediate neuron $c_s$. The activation threshold of $c_{\tilde{pq}}$ through the sigmoidal nonlinearity renders it responsive only when $c_p$ and $c_q$ are activated in a sequential manner. This enables $c_{\tilde{pq}}$ to serve as a sequence detection neuron in a way that is functionally analogous to direction-selective motion detection models proposed for the visual cortex.

\paragraph{Hebbian learning of compositional rules.}
When a sequence neuron $c_{pq}$ fires upon detecting the ordered pair $(r_p,r_q)$, and the target relation neuron $r_k$ is concurrently activated (either by a direct triple or by its inverse through the bidirectional encoding of Eq.~\eqref{eq:encw}), the coincident activity strengthens the synapse between $c_{pq}$ and $r_k$ via the same Oja-like rule used in $L_R$:
\begin{equation}
\Delta w_{k-pq}^{CC} = \delta_{CC}\,c_k\,c_{pq}-\mu_{CC}\,(c_k^{2}+c_{pq}^{2})\,w_{k-pq}^{CC}.
\end{equation}
where the decay term effectively controls for the baseline occurrence rate of relations and chains.

The composition-detection and Hebbian mechanisms described above operate locally on individual triple chains. To discover all compositional rules supported by the data, NSR enumerates every two-hop path in $\mathcal{T}$ and applies the closed-loop check of Eq.~\eqref{eq:comp-rule} together with the weight update of Eq.~\eqref{eq:hebb-c}. Algorithm~\ref{alg:online-comp} implements this global traversal.

\begin{algorithm}[htbp]
\caption{Compositional-rule learning}
\label{alg:online-comp}
\small
\textbf{Notation.} 
$\mathcal{T}$: training triples $(h,r,t)$ with $r\in\{0,\dots,M-1\}$; extended relations $\tilde r\in\{0,\dots,2M-1\}$ encode forward ($\tilde r_k\equiv r_k$) and inverse ($\tilde r_{k+M}\equiv r_k^{-1}$) links via Eq.~\eqref{eq:encw}. 
$c_{pq}$: sequence neuron for $(r_p,r_q)$. 
$c_k$: neuron in $L_C$ representing relation $r_k$. 
$\delta_{CC},\mu_{CC}$: Hebbian rate and decay for $L_C$ synapses.

\vspace{0.5em}
\noindent\begin{minipage}{\linewidth}
\raggedright
1: \textbf{for} each $(h_i,r_p,x_p)\in\mathcal{T}$ \textbf{do}\\
2: \quad\textbf{for} each $(x_p,r_q,x_q)\in\mathcal{T}$ with $r_q\neq r_p^{-1}$ \textbf{do}\\
3: \quad\quad\textbf{for} each $(x_q,r_k,h_i)\in\mathcal{T}$ \textbf{do}\\
4: \quad\quad\quad$c_{pq}\leftarrow$ sequence neuron for $(r_p,r_q)$\\
5: \quad\quad\quad$c_k\leftarrow$ relation neuron for $r_k$\\
6: \quad\quad\quad$\Delta w_{k-pq}^{CC}\leftarrow\delta_{CC}\,c_k\,c_{pq}-\mu_{CC}\,(c_k^2+c_{pq}^2)\,w_{k-pq}^{CC}$\\
7: \quad\quad\quad$w_{k-pq}^{CC}\leftarrow w_{k-pq}^{CC}+\Delta w_{k-pq}^{CC}$\\
8: \quad\quad\textbf{end for}\\
9: \quad\textbf{end for}\\
10: \textbf{end for}
\end{minipage}
\end{algorithm}

\subsubsection{Offline learning on static graphs}
\label{app:offline-pseudo}

On a static training graph, the per-experience updates of Algorithm~\ref{alg:online-comp} admit an exact closed-form evaluation, so the same connectivity $W^{CC}$ can be computed in a single pass over the encoded weights instead of simulating the network dynamics. Two observations underlie this offline variant.

\begin{enumerate}[label=(\roman*), leftmargin=*]
    \item \textbf{Forward episodes reduce to the encoded connectivity.} During composition learning the relation-to-relation couplings $W_{RR}$ are disabled, so an episode initiated at head entity $h$ under relation $\tilde r_i$ returns exactly the encoded neighbors of $h$ under $\tilde r_i$: the dynamics of Eq.~\eqref{eq:fulldyn} reduce to readout through the fixed connectivity of Eq.~\eqref{eq:encw}. Writing $A_{\tilde r}\in\{0,1\}^{N\times N}$ for the adjacency matrix of the encoded triples under $\tilde r$, the two-hop experiences of a chain $(\tilde r_i,\tilde r_j)$ are therefore enumerated by the sparse product $S_{ij}=A_{\tilde r_i}A_{\tilde r_j}$, where $S_{ij}[h,t]$ counts the intermediate entities $x$ with $(h,\tilde r_i,x)$ and $(x,\tilde r_j,t)$ both encoded.
    \item \textbf{Per-experience updates accumulate in closed form.} For an experience $(h,t)$ of the chain, the return readout has the closed form $y_k(h,t)=A_{r_k}[h,t]$: the neuron $c_k$ is active exactly when $(h,r_k,t)$ is encoded. With a running-mean learning rate $\eta_{ij}=1/n_{ij}$, the update of lines~6--7 of Algorithm~\ref{alg:online-comp} is a running average over experiences, so after all $n_{ij}$ experiences of the chain the weight equals the empirical mean
    \begin{equation}
    w_{k-ij}^{CC} \;=\; \frac{\langle A_{r_k},\, S_{ij} \rangle_F}{\sum_{h\neq t} S_{ij}[h,t]},
    \qquad S_{ij}=A_{\tilde r_i}A_{\tilde r_j},
    \label{eq:wcc-fixed-point}
    \end{equation}
    where $\langle\cdot,\cdot\rangle_F$ is the Frobenius inner product and the diagonal of $S_{ij}$ is excluded ($t\neq h$). A constant learning rate $\delta_{CC}$ yields the standard exponentially weighted estimate of the same quantity when triples arrive sequentially.
\end{enumerate}

Algorithm~\ref{alg:offline-comp} evaluates Eq.~\eqref{eq:wcc-fixed-point} directly. It computes the same weights as the online procedure of Algorithm~\ref{alg:online-comp}; no discrete rule set is constructed, stored, or re-grounded at query time. The evidence $n_{ij}$ of each chain is returned together with the weights, and the thresholds $\tau_s$ and $\tau_c$ act only as readout gates during reasoning, in the same way as \texttt{T\_thresh} gates relation readout in \S\ref{sec:reasoning-phase}: a composition neuron $c_{ij}$ participates in reasoning only if $n_{ij}\ge\tau_s$, and a consequent relation $r_k$ is read out from $c_{ij}$ only if $w_{k-ij}^{CC}\ge\tau_c$. The full dynamical implementation and this offline variant attain comparable performance on all benchmarks (\S\ref{sec:EP}).

\begin{algorithm}[H]
\caption{Offline compositional-rule learning}
\label{alg:offline-comp}
\small
\textbf{Notation.} 
$A_{\tilde r}\in\{0,1\}^{N\times N}$: adjacency matrix of encoded triples under extended relation $\tilde r$ (Eq.~\eqref{eq:encw}).
$n_{ij}$: number of experiences (evidence) of chain $(\tilde r_i,\tilde r_j)$.
\textbf{Hyperparameters.} 
$\tau_s$: minimum evidence for $c_{ij}$ to participate in reasoning (query-time gate).
$\tau_c$: minimum weight for reading out a consequent relation (query-time gate).

\vspace{0.5em}
\noindent\begin{minipage}{\linewidth}
\raggedright
1: $W^{CC}\leftarrow 0$\\
2: \textbf{for} each ordered pair $(\tilde r_i,\tilde r_j)$ with $\tilde r_j\neq \tilde r_i^{-1}$ \textbf{do}\\
3: \quad$S_{ij}\leftarrow A_{\tilde r_i}A_{\tilde r_j}$; \quad set diagonal of $S_{ij}$ to zero \textit{// two-hop experiences}\\
4: \quad$n_{ij}\leftarrow\sum_{h,t}S_{ij}[h,t]$\\
5: \quad\textbf{if} $n_{ij}=0$ \textbf{then continue}\\
6: \quad\textbf{for} $k=1,\dots,M$ \textbf{do}\\
7: \quad\quad$w_{k-ij}^{CC}\leftarrow\langle A_{r_k},S_{ij}\rangle_F\,/\,n_{ij}$ \textit{// same weights as Alg.~\ref{alg:online-comp}}\\
8: \quad\textbf{end for}\\
9: \textbf{end for}\\
10: \textbf{return} $W^{CC},\{n_{ij}\}$
\end{minipage}

\end{algorithm}

\section{Details for the Reasoning Phase}
\label{sec:DRP}

\subsection{Reasoning phase (algorithmic form)}
\label{app:reasoning-algorithmic}

Algorithm~\ref{alg:reasoning-phase} restates the five-step procedure of \S\ref{sec:reasoning-phase} in pseudocode. Layer updates follow the dynamics defined in Eq.~\eqref{eq:fulldyn}; the final aggregation follows Eq.~\eqref{eq:score-aggregation}.
\begin{algorithm}[htbp]
\caption{Reasoning phase for query $(h_i,r_k,?)$ with max/sum aggregation.}
\label{alg:reasoning-phase}
\small
\begin{algorithmic}[1]

\REQUIRE Query head $h_i$, query relation $r_k$;
         trained network weights and layer dynamics in
         Eq.~\eqref{eq:fulldyn};
         activation threshold $\texttt{T\_thresh}$;
         gate function $G=\operatorname{TopK}$;
         aggregation mode $a\in\{\mathrm{max},\mathrm{sum}\}$,
         selected on the validation set.
\ENSURE Candidate-tail ranking scores $\mathrm{Score}_a(t)$.

\STATE Initialize path collections
       $\mathcal{P}_R\leftarrow\emptyset$ and
       $\mathcal{P}_C\leftarrow\emptyset$.

\STATE \COMMENT{\textbf{Step 1:} Retrieve equivalent relations from $L_R$}
\STATE Initialize all layer activations
       $X_E,X_Z,X_R,X_C\leftarrow\mathbf{0}$.
\STATE Clamp query relation: $X_R[r_k]\leftarrow 1$.
\STATE Update $L_R$ once using Eq.~\eqref{eq:fulldyn}.
\STATE $\mathcal{R}_{\mathrm{eq}}
       \leftarrow
       \{r_e\mid X_R[r_e]\geq\texttt{T\_thresh}\}$.
\STATE Store $S_R(r_e)\leftarrow X_R[r_e]$
       for each $r_e\in\mathcal{R}_{\mathrm{eq}}$.

\STATE \COMMENT{\textbf{Step 2:} Relational reasoning via $L_Z$}
\STATE Reset $X_E,X_Z\leftarrow\mathbf{0}$;
       retain only $\mathcal{R}_{\mathrm{eq}}$ in $L_R$.
\STATE Activate the head entity: $X_E[e_i]\leftarrow 1$.
\STATE Iterate $L_Z$ and read out $L_E$
       using Eq.~\eqref{eq:fulldyn}.
\STATE Collect the grounded one-hop paths
       $p=(h_i,r_e,t)$ realized by these transitions
       into $\mathcal{P}_R$.
\FOR{each $p=(h_i,r_e,t)\in\mathcal{P}_R$}
    \STATE Assign path-support score $s(p)\leftarrow S_R(r_e)$.
\ENDFOR

\STATE \COMMENT{\textbf{Step 3:} Retrieve relation compositions from $L_C$}
\STATE Re-initialize
       $X_E,X_Z,X_R,X_C\leftarrow\mathbf{0}$.
\STATE Clamp query relation: $X_R[r_k]\leftarrow 1$.
\STATE Propagate activity to $L_C$ and update its dynamics
       using Eq.~\eqref{eq:fulldyn}.
\STATE $\mathcal{C}_{\mathrm{eq}}
       \leftarrow G(X_C)=\operatorname{TopK}(X_C)$.
\STATE Store $S_C(c_e)\leftarrow X_C[c_e]$
       for each $c_e\in\mathcal{C}_{\mathrm{eq}}$.

\STATE \COMMENT{\textbf{Step 4:} Multi-hop reasoning with each composition}
\FOR{each $c_e\in\mathcal{C}_{\mathrm{eq}}$}
    \STATE Reset $X_E,X_Z,X_R,X_C\leftarrow\mathbf{0}$.
    \STATE Activate $c_e$ in $L_C$ and initialize $L_E$ at $e_i$.
    \STATE Decode $c_e$ into the ordered relation chain
           $(r_{p_1},r_{p_2},\ldots,r_{p_\ell})$.
    \FOR{$j=1,\ldots,\ell$}
        \STATE Activate $r_{p_j}$ in $L_R$.
        \STATE Iterate $L_Z$ and read out $L_E$
               using Eq.~\eqref{eq:fulldyn},
               retaining distinct grounded transition traces.
        \STATE Use the reached entities as the starting states
               for the next hop, if any.
    \ENDFOR
    \STATE Collect the completed grounded paths
           $p=(h_i,r_{p_1},x_1,\ldots,r_{p_\ell},t)$
           into $\mathcal{P}_{c_e}$.
    \FOR{each $p\in\mathcal{P}_{c_e}$}
        \STATE Assign path-support score $s(p)\leftarrow S_C(c_e)$.
    \ENDFOR
    \STATE $\mathcal{P}_C
           \leftarrow \mathcal{P}_C\cup\mathcal{P}_{c_e}$.
\ENDFOR

\STATE \COMMENT{\textbf{Step 5:} Max/sum aggregation of path-support scores}
\STATE $\mathcal{P}\leftarrow\mathcal{P}_R\cup\mathcal{P}_C$.
\STATE $\tilde{\mathcal{T}}
       \leftarrow
       \{t\mid \exists p\in\mathcal{P}
       \text{ with terminal entity }t\}$.
\FOR{each $t\in\tilde{\mathcal{T}}$}
    \STATE $\mathcal{P}(t)
           \leftarrow
           \{p\in\mathcal{P}\mid p
           \text{ has terminal entity }t\}$.
    \IF{$a=\mathrm{max}$}
        \STATE $\mathrm{Score}_a(t)
               \leftarrow
               \max_{p\in\mathcal{P}(t)}s(p)$.
    \ELSE
        \STATE $\mathrm{Score}_a(t)
               \leftarrow
               \sum_{p\in\mathcal{P}(t)}s(p)$.
    \ENDIF
\ENDFOR

\RETURN $\mathrm{Score}_a$.

\end{algorithmic}
\end{algorithm}
\subsection{Hyperparameter settings}
\label{app:hyperparams}

Table~\ref{tab:hyperparams} lists the hyperparameters of the simplified variant used in \S\ref{alg:offline-comp}.  They are grouped by the processing stage to which they belong.

\paragraph{Reasoning threshold.} $\texttt{T\_{thresh}}$ is the activation threshold of Step~1 (\S\ref{sec:reasoning-phase}): a relation neuron $r_e$ enters the equivalent-relation set $\mathcal{R}_{\mathrm{eq}}$ only if its activation satisfies $x_e\ge \texttt{T\_{thresh}}$. Given that the Countries dataset contains only two relations, reasoning does not depend on relation similarity; hence, $\texttt{T\_{thresh}}$ is set to infinity.

\paragraph{Learning rate.} The simplified implementation collapses the Hebbian rates $\delta_{RR}$ (Eq.~\eqref{eq:hebb}) and $\delta_{CC}$ (Eq.~\eqref{eq:hebb-c}) to a single scalar $\delta_\theta$, because the simplified variant performs one-pass counting instead of iterative Oja updates.  The value reported in Table~\ref{tab:hyperparams} therefore serves as the effective learning rate for both $W_{RR}$ and $W_{CC}$.

\paragraph{Compositional-rule thresholds.} $\tau_s$ and $\tau_c$ are the query-time readout gates of Algorithm~\ref{alg:offline-comp}: during reasoning, a composition neuron $c_{ij}$ participates only if its evidence $n_{ij}$ is at least $\tau_s$, and a consequent relation is read out from $c_{ij}$ only if its weight $w_{k-ij}^{CC}$ is at least $\tau_c$. All values were obtained with \texttt{Optuna}~\cite{akiba2019optuna}; the search ranges were $\texttt{T\_{thresh}}\in[0.05,1.0]$, $\delta_\theta\in[0.1,1.0]$, $\tau_s\in[1,50]$, and $\tau_c\in[0.05,0.5]$.  NSR is not highly sensitive to these settings.

\begin{table}[h]
\centering
\small
\setlength{\tabcolsep}{4pt}
\caption{Hyperparameters for the main KG benchmarks}
\label{tab:hyperparams}
\begin{tabular}{@{}l cccc@{}}
\toprule
\textbf{Dataset} & $\texttt{T\_{thresh}}$ & $\tau_s$ & $\tau_c$ & $\delta_\theta$ \\
\midrule
Nations & 0.22 & 23 & 0.20 & 0.52 \\
Kinship & 0.90 & 27 & 0.17 & 0.20 \\
Countries S3 & $\inf$ & 1 & 1.00 & 1.00 \\
Kinship1990\_EXTENDED & 0.46 & 23 & 0.13 & 0.34 \\
\bottomrule
\end{tabular}
\end{table}

\section{Datasets and Detailed Experiments}
\label{app:datasets-experiments}

\subsection{Datasets}
\label{app:datasets}

We evaluate NSR on knowledge-graph datasets of varying scale and relational complexity. Nations, Kinship, and Countries S3 are standard benchmarks in the KG-embedding literature. Kinship1990\_EXTENDED is an extended derivative of the classical Kinship domain that we introduce to stress-test compositional relational reasoning; its construction is detailed in \S~\ref{app:kinship-extended}. WN18RR, FB15k-237, and YAGO3-10 are standard large-scale benchmarks. Table~\ref{tab:dataset_stats} summarizes the number of entities, relation types, and triples (edges) in each split.

\begin{table}[htbp]
\centering
\small
\setlength{\tabcolsep}{6pt}
\caption{Statistics of the datasets used in our experiments. All counts in the last five columns refer to the number of edges (triples).}
\label{tab:dataset_stats}
\begin{tabular}{@{}lrrrrrr@{}}
\toprule
\textbf{Dataset} & \textbf{Entities} & \textbf{Relations} & \textbf{Total} & \textbf{Train} & \textbf{Val} & \textbf{Test} \\
\midrule
Nations & 14 & 55 & 1,992 & 1,592 & 199 & 201 \\
Kinship & 104 & 26 & 10,686 & 8,544 & 1,068 & 1,074 \\
Kinship1990\_EXTENDED & 480 & 14 & 2,240 & 1,568 & 224 & 448 \\
Countries\_S3 & 271 & 2 & 1,033 & 985 & 24 & 24 \\
WN18RR & 40,943 & 11 & 93,003 & 86,835 & 3,034 & 3,134 \\
FB15k-237 & 14,541 & 237 & 310,116 & 272,115 & 17,535 & 20,466 \\
YAGO3-10 & 123,182 & 37 & 1,089,040 & 1,079,040 & 5,000 & 5,000 \\
\bottomrule
\end{tabular}
\end{table}

\subsection{How to Construct Kinship1990\_EXTENDED}
\label{app:kinship-extended}

\paragraph{Original data and its limitations.} The original Kinship dataset \cite{917563} contains 112 triples over 24 persons from two isomorphic families, with 12 binary relations: \texttt{wife}, \texttt{husband}, \texttt{mother}, \texttt{father}, \texttt{daughter}, \texttt{son}, \texttt{sister}, \texttt{brother}, \texttt{aunt}, \texttt{uncle}, \texttt{niece}, and \texttt{nephew}. Three features limit its utility for testing compositional reasoning: (i) \textit{No symmetric relations}---every relation is directional, so link-prediction can be solved by learning inverses alone; (ii) \textit{Flat relation labels}---genealogical concepts such as \texttt{aunt} or \texttt{uncle} are provided as atomic symbols rather than explicit compositions of primitives (e.g., \texttt{mother} $\circ$ \texttt{sister}), so models are not required to discover relational grammar; (iii) \textit{Small scale}---112 triples provide insufficient coverage to stress-test generalization.

\paragraph{Reconstructing the base graph.} Because the two families have identical structure, the complete genealogical tree is fully determined by the original triples. We first reconstruct this tree, obtaining all parent--child and spousal links.

\paragraph{Injecting symmetric relations.} We add \texttt{sibling} as an explicitly symmetric relation. For every pair of distinct children sharing at least one parent in the reconstructed tree, we insert \texttt{sibling}$(X,Y)$ and \texttt{sibling}$(Y,X)$. Unlike the original gender-specific \texttt{brother} and \texttt{sister}, \texttt{sibling} forces the model to respect an equivalence constraint that cannot be reduced to a directional inverse.

\paragraph{Extracting compositional chain relations.} We define five new relation types by enumerating multi-hop genealogical paths in the reconstructed tree and abstracting them into explicit labels:
\begin{itemize}
    \item \texttt{grandmother\_chain}: the mother of one's father or mother;
    \item \texttt{grandfather\_chain}: the father of one's father or mother;
    \item \texttt{maternal\_aunt\_chain}: the sister of one's mother;
    \item \texttt{paternal\_uncle\_chain}: the brother of one's father;
    \item \texttt{cousin\_chain}: the child of one's parent's sibling.
\end{itemize}
For example, \texttt{cousin\_chain}$(h,t)$ is generated whenever there exist intermediates $x,y$ such that $x$ is a parent of $h$, $y$ is a sibling of $x$, and $t$ is a child of $y$. These chain relations are not redundant with the original flat labels; they make compositional substructure explicit and therefore require the model to reuse primitive relations rather than memorize atomic mappings.

\paragraph{Statistics and splits.} The extended graph contains 480 entities, 14 relation types, and 2,240 triples. We follow the protocol: 70\% train, 10\% validation, 20\% test. Because the underlying family tree is small and structurally deterministic, all base-relation edges necessary for genealogical consistency appear in the training split; held-out test triples are drawn predominantly from the extended relation set so that accurate prediction benefits from compositional reuse of the trained primitives.

\subsection{Detailed Experiments}
\label{app:detailed-experiments}

\paragraph{Embedding-Based  baseline implementations.} 
All neural baselines are trained with PyKEEN~v1.10 \cite{ali2021pykeen} under identical data splits. 
Table~\ref{tab:baseline-hparams} lists the architecture and optimization settings shared across embedding models. 
The embedding dimension is set to $100$ for small-scale datasets (e.g., Nations and Kinship) and to $200$ for large-scale datasets (e.g., WN18RR, FB15k-237 and YAGO3-10). 
DistMult, ComplEx, RESCAL, and ConvE use the LCWA training loop; TransE and RotatE use sLCWA. 
ConvE requires inverse triples (create\_inverse=True) per PyKEEN's implementation and uses $32$ output channels with dropout rates $\{0.2,0.2,0.3\}$. 
The optimizer is Adam with learning rate $10^{-3}$ for all models except RotatE ($5\times10^{-4}$). 
Training runs for at most $150$ epochs with batch size $32$.

\begin{table}[h]
\centering
\small
\setlength{\tabcolsep}{4pt}
\caption{Neural baseline hyperparameters and training configuration.}
\label{tab:baseline-hparams}
\begin{tabular}{@{}lcccccc@{}}
\toprule
\textbf{Model} & \textbf{Dim} & \textbf{Loop} & \textbf{Inverse} & \textbf{LR} & \textbf{Epochs} & \textbf{Special} \\
\midrule
TransE   & 100 & sLCWA & No  & $1\mathrm{e}{-3}$ & 150 & scoring\_fct\_norm=1 \\
DistMult & 100 & LCWA  & No  & $1\mathrm{e}{-3}$ & 150 & --- \\
ComplEx  & 100 & LCWA  & No  & $1\mathrm{e}{-3}$ & 150 & --- \\
RotatE   & 100 & sLCWA & No  & $5\mathrm{e}{-4}$ & 150 & --- \\
ConvE    & 100,200 & LCWA  & Yes & $1\mathrm{e}{-3}$ & 150 & out\_ch=32, dropouts $\{0.2,0.2,0.3\}$ \\
RESCAL   & 100 & LCWA  & No  & $1\mathrm{e}{-3}$ & 150 & --- \\
\bottomrule
\end{tabular}
\end{table}

\paragraph{Rule Learning baseline implementations.} 
RNNLogic operates \emph{without pretrained knowledge-graph embeddings}. 
It first mines relational paths up to length $3$ using the C++ miner ($16$ threads), then learns rule weights via a lightweight predictor trained for $10$ iterations (Adam, learning rate $5\times10^{-3}$, weight decay $0$, hidden dimension $32$, batch size $16$). 
Label smoothing ($0.1$) is applied during training, and expectation-based ranking (\texttt{expectation=True}) is used at evaluation. 
Unlike neural baselines, no entity or relation embeddings are supplied or learned.

\paragraph{Reproducibility.} We used five random seeds: [42, 43, 44, 45, 46], and the performance of these baseline models and NSR is reported in \S~\ref{sec:EP}. PyKEEN handles its own internal seeding for negative sampling and parameter initialization. NSR and all baselines are evaluated on identical train/validation/test splits for each dataset.

\begin{table}[h]
\centering
\small
\setlength{\tabcolsep}{2pt}
\caption{Symbolic baseline hyperparameters and training configuration.}
\label{tab:symbolic-hparams}
\begin{tabular}{@{}>{\raggedright\arraybackslash}p{3cm}>{\centering\arraybackslash}p{1.2cm}>{\centering\arraybackslash}p{1.3cm}>{\centering\arraybackslash}p{1.3cm}>{\centering\arraybackslash}p{1.3cm}>{\raggedright\arraybackslash}p{5cm}@{}}
\toprule
\textbf{Model} & \textbf{Rule Len} & \textbf{Hidden Dim} & \textbf{LR} & \textbf{Iters} & \textbf{Special} \\
\midrule
RNNLogic (Miner)    & 3           & --- & --- & --- & 16 threads \\
RNNLogic (Predictor)& ---         & 32  & $5\mathrm{e}{-3}$ & 10 & batch=16, smoothing=0.1, expectation=True, \textbf{without embedding} \\
\bottomrule
\end{tabular}
\end{table}

\subsection{Additional Benchmark Results}
\label{app:additional-benchmarks}

We report three complementary evaluations deferred from the main text: Countries S3, which probes multi-hop chain generalization; Kinship1990\_EXTENDED, our controlled compositional stress test (\S\ref{app:kinship-extended}); and WN18RR, a standard large-scale benchmark. We then compare NSR against eight symbolic, neuro-symbolic, and neural reasoning baselines across all five benchmarks, and report training and inference costs.

\paragraph{Countries S3.} Test queries ask for a country's continent, but training triples only link neighboring countries and upward \texttt{locatedIn} edges; no country--continent fact is observed directly, so answering requires composing \texttt{neighborOf} and \texttt{locatedIn} chains. Table~\ref{tab:countries_results} shows that NSR solves the task perfectly, while embedding baselines collapse (best MRR $0.199$) because static vector similarity cannot recover unobserved multi-step paths, and RNNLogic reaches only $0.341$ because externally mined templates miss the compositional regularity.

\begin{table}[htbp]
\centering
\footnotesize
\setlength{\tabcolsep}{4pt}
\caption{Performance comparison on Countries S3. Best results in each column are highlighted in bold. Hits@$k$ are reported as percentage values in $[0,100]$; standard deviations are in units of $10^{-2}$ (MRR) and percentage points (Hits@$k$).}
\label{tab:countries_results}
\begin{tabular}{@{}l ccc c@{}}
\toprule
\textbf{Model} & MRR $\uparrow$ & Hits@1 $\uparrow$ & Hits@3 $\uparrow$ & \textbf{Train Time (s)} \\
\midrule
ConvE    & 0.1987{\tiny$\pm$2.47} & 11.25{\tiny$\pm$2.50} & 19.17{\tiny$\pm$4.04} & 20.3{\tiny$\pm$2.7} \\
DistMult & 0.1761{\tiny$\pm$1.42} & 7.50{\tiny$\pm$1.02} & 20.42{\tiny$\pm$3.06} & 17.4{\tiny$\pm$1.0} \\
RotatE   & 0.1184{\tiny$\pm$2.42} & 4.17{\tiny$\pm$1.86} & 10.00{\tiny$\pm$4.45} & 19.4{\tiny$\pm$2.9} \\
RESCAL   & 0.1906{\tiny$\pm$1.00} & 10.42{\tiny$\pm$2.64} & 18.33{\tiny$\pm$2.04} & 14.7{\tiny$\pm$2.1} \\
ComplEx  & 0.0290{\tiny$\pm$2.68} & 0.83{\tiny$\pm$1.67} & 2.08{\tiny$\pm$4.17} & 18.4{\tiny$\pm$2.2} \\
TransE   & 0.1182{\tiny$\pm$0.61} & 0.00{\tiny$\pm$0.00} & 14.17{\tiny$\pm$2.76} & 18.9{\tiny$\pm$1.7} \\
RNNLogic & 0.3409{\tiny$\pm$3.29} & 0.00{\tiny$\pm$0.00} & 62.50{\tiny$\pm$12.36} & 5.1{\tiny$\pm$0.0} \\
\midrule
NSR      & \textbf{1.0000}{\tiny$\pm$0.00} & \textbf{100.00}{\tiny$\pm$0.00} & \textbf{100.00}{\tiny$\pm$0.00} & \textbf{0.0012}{\tiny$\pm$0.0004} \\
\bottomrule
\end{tabular}
\end{table}

\paragraph{Kinship1990\_EXTENDED.} On the composition-focused extension, test triples are drawn predominantly from the five composition-defined chain relations (\S\ref{app:kinship-extended}). NSR attains the best MRR ($0.953$) and Hits@1 ($94.3$), surpassing all baselines including DistMult, which is strong on MRR ($0.941$) but relies on memorized symmetric structure; the inversion relative to classical Kinship, where ConvE leads, is discussed in \S\ref{sec:EP}.

\begin{table}[htbp]
\centering
\footnotesize
\setlength{\tabcolsep}{4pt}
\caption{Performance comparison on Kinship1990\_EXTENDED. Conventions as in Table~\ref{tab:countries_results}.}
\label{tab:kinship_ext_results}
\begin{tabular}{@{}l ccc c@{}}
\toprule
\textbf{Model} & MRR $\uparrow$ & Hits@1 $\uparrow$ & Hits@3 $\uparrow$ & \textbf{Train Time (s)} \\
\midrule
ConvE    & 0.8265{\tiny$\pm$0.91} & 74.93{\tiny$\pm$1.47} & 87.83{\tiny$\pm$0.72} & 60.5{\tiny$\pm$7.5} \\
DistMult & 0.9409{\tiny$\pm$0.15} & 90.62{\tiny$\pm$0.30} & \textbf{97.75}{\tiny$\pm$0.25} & 31.9{\tiny$\pm$4.0} \\
RotatE   & 0.5988{\tiny$\pm$4.47} & 51.94{\tiny$\pm$5.27} & 64.40{\tiny$\pm$4.20} & 26.0{\tiny$\pm$1.1} \\
RESCAL   & 0.0144{\tiny$\pm$0.28} & 0.22{\tiny$\pm$0.12} & 0.80{\tiny$\pm$0.31} & 30.3{\tiny$\pm$5.3} \\
ComplEx  & 0.0220{\tiny$\pm$0.34} & 0.42{\tiny$\pm$0.23} & 1.41{\tiny$\pm$0.50} & 31.4{\tiny$\pm$1.6} \\
TransE   & 0.2249{\tiny$\pm$0.58} & 2.19{\tiny$\pm$0.69} & 33.42{\tiny$\pm$1.58} & 25.3{\tiny$\pm$2.4} \\
RNNLogic & 0.8107{\tiny$\pm$1.82} & 80.45{\tiny$\pm$1.92} & 81.38{\tiny$\pm$1.78} & 9.3{\tiny$\pm$1.4} \\
\midrule
NSR      & \textbf{0.9533}{\tiny$\pm$0.30} & \textbf{94.33}{\tiny$\pm$0.48} & 96.47{\tiny$\pm$0.09} & \textbf{0.0451}{\tiny$\pm$0.0002} \\
\bottomrule
\end{tabular}
\end{table}

\paragraph{WN18RR.} Table~\ref{tab:wn18rr_results} reports full filtered metrics on WN18RR, together with the symbolic and neuro-symbolic baselines (MRR; see Table~\ref{tab:kg-benchmarks} for the remaining datasets). NSR attains the best Hits@1 ($0.463$) among all evaluated methods and is competitive with the strongest embedding models on MRR, while the GNN reasoner NBFNet remains stronger; we view NSR as competitive rather than uniformly superior on this benchmark.

\begin{table}[htbp]
\centering
\footnotesize
\setlength{\tabcolsep}{4pt}
\caption{Filtered link prediction on WN18RR. Symbolic and neuro-symbolic baselines report MRR only; ``---'' indicates unreported or not scalable. Best result in each column in bold.}
\label{tab:wn18rr_results}
\begin{tabular}{@{}l cccc@{}}
\toprule
\textbf{Method} & MRR & Hits@1 & Hits@3 & Hits@10 \\
\midrule
TransE   & .226 & --- & --- & .501 \\
DistMult & .430 & .390 & .440 & .490 \\
ConvE    & .430 & .400 & .440 & .520 \\
ComplEx  & .440 & .410 & .460 & .510 \\
RotatE   & .476 & .428 & .492 & .571 \\
BoxE     & .451 & .400 & .472 & .541 \\
ModE     & .472 & .427 & .486 & .564 \\
HAKE     & .497 & .452 & \textbf{.516} & .582 \\
SectorE  & .475 & .421 & .478 & \textbf{.586} \\
\midrule
AnyBURL        & .5658 & --- & --- & --- \\
AMIE           & .4157 & --- & --- & --- \\
PRA / PathRank & .0556 & --- & --- & --- \\
NTP            & ---   & --- & --- & --- \\
NeuralLP       & .4677 & --- & --- & --- \\
NCRL           & .4070 & --- & --- & --- \\
MINERVA        & .4890 & --- & --- & --- \\
NBFNet         & \textbf{.5976} & --- & --- & --- \\
\midrule
NSR & .472 & \textbf{.463} & .483 & .484 \\
\bottomrule
\end{tabular}
\end{table}

\paragraph{Comparison with symbolic and neuro-symbolic reasoners.} Table~\ref{tab:kg-benchmarks} compares NSR against eight reasoning baselines spanning symbolic rule mining (AnyBURL, AMIE), path ranking (PRA/PathRank), differentiable rule learning (NTP, NeuralLP, NCRL), RL-based path search (MINERVA), and GNN reasoning (NBFNet) on all five benchmarks. NSR outperforms the evaluated differentiable rule learners on every dataset they complete, performs comparably to strong symbolic rule miners overall, and achieves the best result on YAGO3-10, while NBFNet remains stronger on WN18RR and FB15k-237 and AMIE is strongest on Nations. Because the three large benchmarks are heterogeneous and not designed around NSR's compositional structure, these results indicate that NSR's performance is not confined to small, composition-aligned benchmarks.

\begin{table}[htbp]
\centering
\caption{Link prediction results (MRR) on all KG completion benchmarks. Best results are in \textbf{bold}; asterisks (*) denote values taken from prior work or official repositories, and ``---'' indicates results not reported or the method does not scale to the dataset.}
\label{tab:kg-benchmarks}
\small
\setlength{\tabcolsep}{4pt}
\begin{tabular}{@{}l ccccc@{}}
\toprule
\textbf{Method} & \textbf{Nations} & \textbf{Kinship} & \textbf{WN18RR} & \textbf{YAGO3-10} & \textbf{FB15k-237} \\
\midrule
AnyBURL        & 0.7994 & 0.6768 & 0.5658 & 0.5589 & 0.332$^*$ \\
AMIE           & \textbf{0.8559} & 0.6767 & 0.4157 & 0.5473 & 0.2170 \\
PRA / PathRank & 0.5933 & 0.6296 & 0.0556 & 0.4678 & 0.0972 \\
NTP            & 0.6223 & 0.612$^*$ & --- & --- & --- \\
NeuralLP       & 0.6841 & 0.6072 & 0.4677 & --- & 0.3166 \\
NCRL           & 0.4571 & 0.6050 & 0.4070 & 0.380$^*$ & 0.300$^*$ \\
MINERVA        & 0.5865 & 0.6253 & 0.4890 & --- & 0.2734 \\
NBFNet         & 0.7479 & \textbf{0.7445} & \textbf{0.5976} & 0.4946 & \textbf{0.5114} \\
NSR            & 0.8142 & 0.6515 & 0.4720 & \textbf{0.5893} & 0.3649 \\
\bottomrule
\end{tabular}
\end{table}

\paragraph{Training and inference cost.} Table~\ref{tab:efficiency} reports end-to-end training time and full-test-set inference time. NSR trains in seconds to minutes across all benchmarks---orders of magnitude below iterative neural training on the large datasets---while inference cost remains modest; measurements on baselines span different hardware and stopping criteria, so we report them as indicative rather than as strict multiplicative speedups.

\begin{table}[htbp]
\centering
\caption{Training time / full-test-set inference time (seconds unless noted). $^{**}$quoted from the official AnyBURL-23-1 release; other local runs use the same machine. NBFNet entries use official training profiles, so cross-hardware comparisons are indicative. OOM: out of memory.}
\label{tab:efficiency}
\small
\setlength{\tabcolsep}{4pt}
\begin{tabular}{@{}l ccccc@{}}
\toprule
\textbf{Method} & \textbf{Nations} & \textbf{Kinship} & \textbf{WN18RR} & \textbf{YAGO3-10} & \textbf{FB15k-237} \\
\midrule
AnyBURL        & 64\,s / 16\,s   & 63\,s / 36\,s    & 603\,s / 11\,s   & 1000\,s / 424\,s & 1000\,s$^{**}$ / --- \\
AMIE           & 2246\,s / $<$0.1\,s & 2\,s / 0.3\,s & 2\,s / $<$0.1\,s & 83\,s / 0.6\,s  & 8\,s / 1.2\,s \\
PRA / PathRank & 34\,s / 0.5\,s  & 86\,s / 0.3\,s   & 22\,s / 156\,s   & 239\,s / 354\,s  & 229\,s / 841\,s \\
NTP            & 1.9\,h / 8\,s   & ---              & ---              & ---              & --- \\
NeuralLP       & 40\,s / 8\,s    & 26\,s / 4\,s     & 1.1\,h / 129\,s  & --- (OOM)        & 11.9\,h / 533\,s \\
NCRL           & 134\,s / 43\,s  & 99\,s / 7\,s     & 244\,s / 24\,s   & ---              & --- \\
MINERVA        & 3160\,s / 8\,s  & 1.2\,h / 13\,s   & 3416\,s / 44\,s  & ---              & 3.3\,h / 333\,s \\
NBFNet         & 29\,s / 0.1\,s  & 77\,s / 0.3\,s   & 5.6\,h / 12\,s   & 2.3\,h / 422\,s  & 10.4\,h / 46\,s \\
\midrule
NSR            & 3\,s / 0.4\,s   & 11\,s / 1.2\,s   & 3.3\,s / 32\,s   & 1142\,s / 107\,s & 2019\,s / 55\,s \\
\bottomrule
\end{tabular}
\end{table}

\subsection{Ablation Study}
\label{app:ablation}

We ablate the four learnable or structural components of NSR to isolate their individual contributions: inverse-relation encoding (Eq.~\eqref{eq:encw}), relation-equivalence retrieval (Step~1 of \S\ref{sec:reasoning-phase}), compositional inference (Steps~3--4 of \S\ref{sec:reasoning-phase}), and Hebbian learning (Eqs.~\eqref{eq:hebb} and~\eqref{eq:hebb-c}). All ablations are conducted with the full dynamical model on Nations, Kinship, and Kinship1990\_EXTENDED, under the filtered link-prediction protocol of \S\ref{app:exp_setup}. Each variant is evaluated on five paired seeds; we report MRR (mean $\pm$ population standard deviation across seeds) together with $\Delta$MRR relative to the paired full model. Ties between equally scored entities are broken by seed-dependent uniform noise of magnitude $10^{-6}$.

\begin{table}[htbp]
\centering
\footnotesize
\setlength{\tabcolsep}{4pt}
\caption{Component-wise ablations of NSR (MRR, mean $\pm$ std over five seeds). Values in parentheses denote $\Delta$MRR relative to the paired full model. Best results in each column are highlighted in bold.}
\label{tab:ablation}
\begin{tabular}{@{}l ccc@{}}
\toprule
\textbf{Variant} & \textbf{Nations} & \textbf{Kinship} & \textbf{Kinship1990\_EXTENDED} \\
\midrule
Full model
  & \textbf{0.81}{\tiny$\pm$0.03}
  & \textbf{0.65}{\tiny$\pm$0.01}
  & \textbf{0.95}{\tiny$\pm$0.00} \\
\quad w/o inverse encoding
  & 0.66{\tiny$\pm$0.03} ($-$0.15)
  & 0.05{\tiny$\pm$0.00} ($-$0.60)
  & 0.10{\tiny$\pm$0.00} ($-$0.85) \\
\quad w/o relation-equivalence retrieval
  & 0.61{\tiny$\pm$0.01} ($-$0.20)
  & 0.42{\tiny$\pm$0.00} ($-$0.24)
  & 0.82{\tiny$\pm$0.00} ($-$0.12) \\
\quad w/o compositional inference
  & 0.79{\tiny$\pm$0.03} ($-$0.03)
  & 0.48{\tiny$\pm$0.01} ($-$0.17)
  & 0.46{\tiny$\pm$0.00} ($-$0.48) \\
\quad w/o Hebbian learning
  & 0.36{\tiny$\pm$0.01} ($-$0.45)
  & 0.05{\tiny$\pm$0.00} ($-$0.60)
  & 0.02{\tiny$\pm$0.00} ($-$0.93) \\
\bottomrule
\end{tabular}
\end{table}

The variants are defined as follows. \emph{W/o inverse encoding}: the model is retrained from scratch with the inverse half of the encoding in Eq.~\eqref{eq:encw} disabled---inverse triples $(t,r^{-1},h)$ are never encoded and the inverse relation neurons $\{r_{i+M}\}$ of $L_R$ are unused; both learning phases are otherwise unchanged. \emph{W/o relation-equivalence retrieval}: training is identical to the full model and the same trained network is reused; only the reasoning phase is modified, with the $W_{RR}$ propagation of Step~1 in \S\ref{sec:reasoning-phase} disabled at query time, so a query is answered using only the query relation itself and its associated compositions. \emph{W/o compositional inference}: again the same trained network as the full model is reused, and the $L_C$ pathway (Steps~3--4 of \S\ref{sec:reasoning-phase}) is disabled at query time, so candidate tails come exclusively from equivalent-relation paths $\tilde{\mathcal{T}}^{(1)}$. \emph{W/o Hebbian learning}: the encoding phase (Eq.~\eqref{eq:encw}) proceeds as in the full model, but both Hebbian learning phases are skipped, so $W_{RR}$ and $W^{CC}$ remain at their zero initialization; the reasoning pipeline is run unchanged.

\paragraph{Findings.} The degradation pattern is functionally specific rather than uniform, showing that each component serves a distinct role. Inverse encoding is decisive on the kinship benchmarks ($-0.60$ and $-0.85$ MRR): directed kinship relations such as \texttt{son\_of} or \texttt{grandmother\_chain} can only be traversed through their inverses, whereas Nations ($-0.15$) usually offers an equivalent forward relation as a detour. Relation-equivalence retrieval contributes consistently across all datasets ($-0.12$ to $-0.24$), confirming that the learned $W_{RR}$ couplings provide a reliable first source of candidate relations at query time. Compositional inference matters most on Kinship1990\_EXTENDED ($-0.48$), whose test queries require unseen multi-hop compositions by construction, and least on Nations ($-0.03$), where most test queries are solvable by single-hop equivalence---mirroring the intended difference between the two benchmarks. Finally, disabling Hebbian learning causes the largest overall degradation ($-0.45$, $-0.60$, $-0.93$), showing that the learned associative connectivity is essential to performance rather than a merely decorative biological motif. The small residual MRR of this variant is expected: with $W_{RR}$ and $W^{CC}$ at zero, no equivalent relation or composition can be retrieved, entity scores degenerate, and ranks are effectively determined by tie-breaking noise, whose expected reciprocal rank is nonzero on small entity sets.

We stress the scope of this evidence: these ablations validate the \emph{computational} role of each implemented component. They are not evidence for or against the biological realism of the corresponding mechanisms, which we present as computational inspiration.

\paragraph{Aggregation rule.} We also ablate the max aggregation of Eq.~\eqref{eq:score-aggregation} against sum aggregation. The two rules exhibit a genuine trade-off: max aggregation can be dominated by a single strong but spurious path, whereas sum aggregation pools convergent evidence but may overcount correlated paths. For example, on the Kinship query (\texttt{person80}, \texttt{term16}, $?$), the target \texttt{person25} has no direct evidence; under max aggregation a competitor (\texttt{person32}, score $0.662$) slightly outscores the target ($0.641$), while under sum aggregation the target accumulates support from $132$ grounded paths (total $46.23$, versus $1.55$ from $3$ paths for the competitor) and becomes the unique top-ranked answer. We therefore select the aggregation mode on validation data and treat NSR's outputs as ranking scores rather than calibrated probabilities.

\section{Additional Analyses and Reasoning Traces}
\label{app:additional-analyses}

\subsection{Symmetry Reasoning Trace}
\label{app:additional-traces}

\begin{figure}[htbp]
    \centering
    \includegraphics[width=1\linewidth]{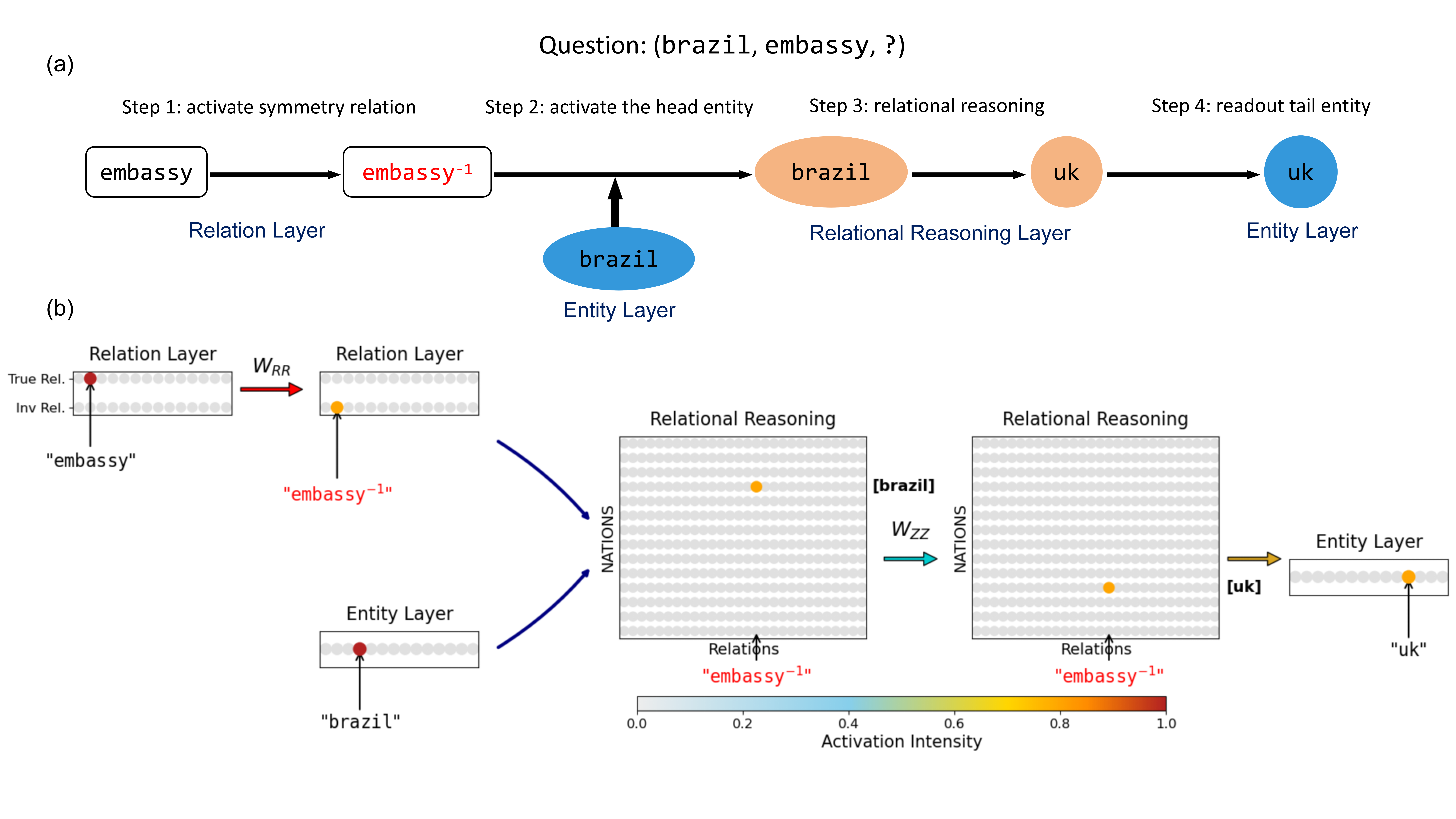}
\caption{\textbf{Symmetry-reasoning trace (symmetry branch only).} 
In Path~A, symmetry and inverse evaluations are performed in parallel. The complete trace therefore contains two branches: a symmetry branch that retains the original relation neuron, and an inverse branch that activates the inverse-relation neuron. For visual clarity, only the symmetry branch is shown here; the inverse branch follows the same gated-diffusion and binding structure with relation inversion.}
\label{fig:symmetry-trace}
\end{figure}

Figure~\ref{fig:symmetry-trace} illustrates the activation trajectory for symmetry reasoning on the query \texttt{(Brazil, embassy, ?)}. 
During the associative phase, the query relation \texttt{embassy} diffuses through $W_{RR}$, and the complementary gate promotes both symmetric and inverse counterparts in parallel. 
The full inference thus comprises two structurally isomorphic branches: one retaining the original relation \texttt{embassy} (symmetry) and one activating \texttt{embassy}$^{-1}$ (inverse). 
Because the two branches differ only in whether the relation neuron is inverted, we display only the symmetry branch for clarity; the inverse branch can be obtained by following the same pipeline with the $W_{RR}$ dynamics.

\subsection{Latent Relational Structure of Countries}
\label{sec:latent_countries}

The Countries dataset is built from only two primitive relations,
\texttt{locatedIn} and \texttt{neighborOf}, distributed over three entity strata:
countries, regions, and continents. Crucially, \texttt{neighborOf} edges are
observed \emph{only} between country-level entities; the dataset contains no
direct adjacency between regions or between continents, even though such
adjacencies are clearly entailed by the underlying geography. Recovering these
missing higher-order neighborhoods therefore provides a natural probe of NSR's
compositional generalization, requiring the model to abstract the same
adjacency relation across the entity hierarchy rather than within a fixed layer.

Two chained rules suffice to lift adjacency upward across strata:
\begin{align}
\texttt{nb}(c_a, c_b) \wedge \texttt{loc}(c_a, R_a) \wedge \texttt{loc}(c_b, R_b)
&\Rightarrow \texttt{nb}(R_a, R_b), \label{eq:country_to_region}\\
\texttt{nb}(R_a, R_b) \wedge \texttt{loc}(R_a, K_a) \wedge \texttt{loc}(R_b, K_b)
&\Rightarrow \texttt{nb}(K_a, K_b), \label{eq:region_to_continent}
\end{align}
where \texttt{nb}\,$\equiv$\,\texttt{neighborOf} and \texttt{loc}\,$\equiv$\,\texttt{locatedIn}; $c_\bullet$, $R_\bullet$, and $K_\bullet$ index countries, regions, and
continents respectively.

\begin{figure}[h]
  \centering
  \includegraphics[width=0.95\linewidth]{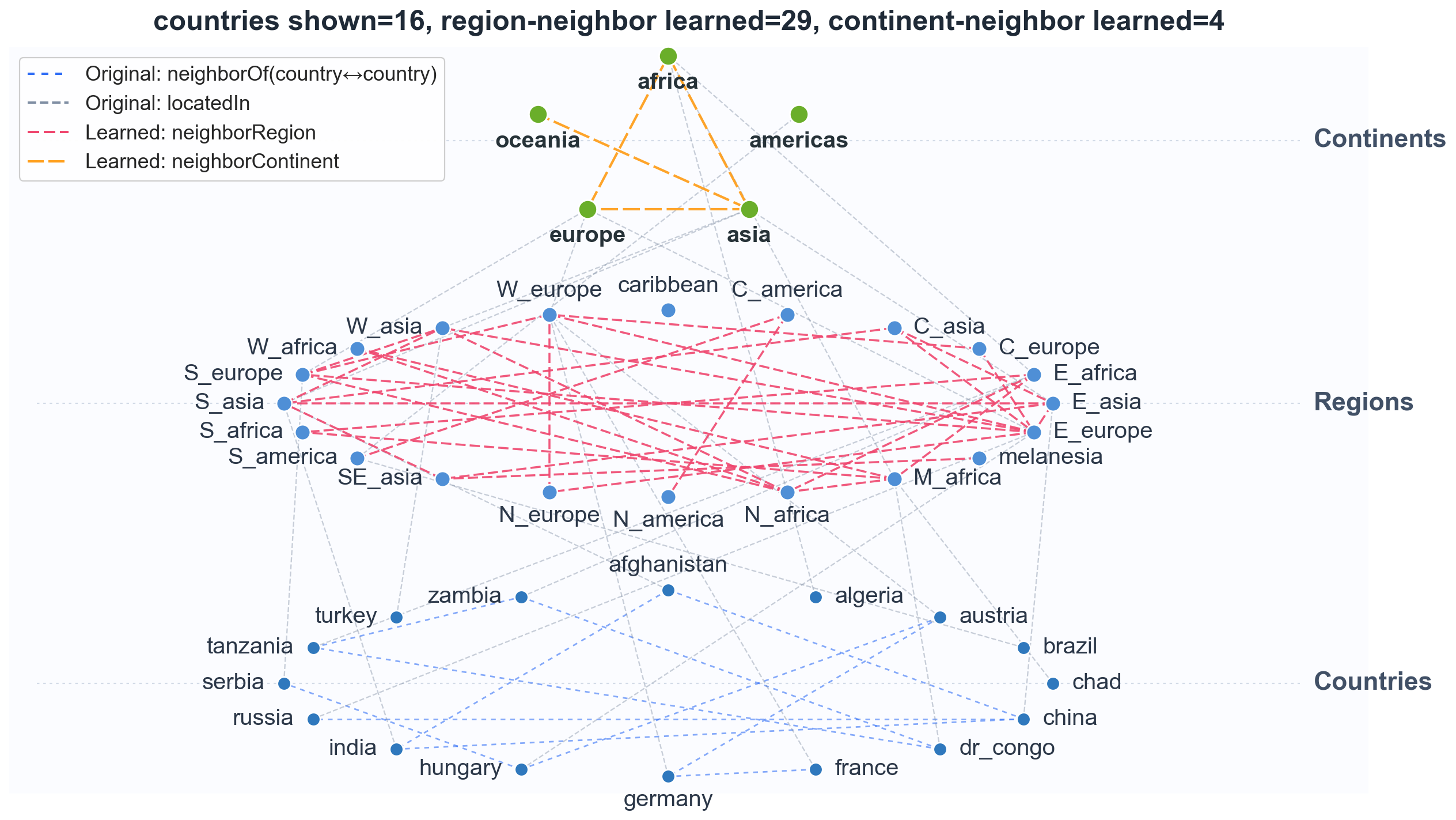}
  \caption{\textbf{Discovery of latent \texttt{neighborOf} on
  Countries.} Blue and gray dashed edges denote the original
  \texttt{neighborOf} and \texttt{locatedIn} facts; red and orange dashed
  edges denote the region- and continent-level adjacencies induced by NSR.Region prefixes abbreviate cardinal directions: N (northern), S (southern), E (eastern), W (western), C (central), M (middle), SE (southeastern).
  Only the $16$ most active countries are shown; all regions and continents
  are displayed in full.}
  \label{fig:countries_layered}
\end{figure}

Figure~\ref{fig:countries_layered} renders the dataset as a three-tier
layered graph. Without ever observing a single direct edge of either type during training, NSR recovers $29$ region-neighbor
and $4$ continent-neighbor edges, and the induced adjacencies are
geographically coherent: the continent layer assembles the expected
Africa--Europe--Asia triangle together with an Asia--Oceania link, while
the region layer reproduces dense intra-Europe, intra-Africa, and intra-Asia
neighborhoods (e.g., \texttt{western\_europe} adjoining
\texttt{northern\_europe} and \texttt{southern\_europe}).

\end{document}